\pdfoutput=1
\documentclass[10pt]{article} % For LaTeX2e
\usepackage[accepted]{tmlr}
\usepackage{amsmath,amsfonts,bm}

\def\eqref#1{equation~\ref{#1}}
\def\1{\bm{1}}

\DeclareMathAlphabet{\mathsfit}{\encodingdefault}{\sfdefault}{m}{sl}
\SetMathAlphabet{\mathsfit}{bold}{\encodingdefault}{\sfdefault}{bx}{n}

\usepackage[colorlinks=true, hidelinks]{hyperref}
\usepackage{url}

\usepackage{subcaption}
\usepackage{adjustbox}
\usepackage{siunitx}
\usepackage{scalefnt}
\usepackage{wrapfig}
\usepackage{tikz}
\usepackage{booktabs}
\usepackage{mathtools}
\usepackage{multirow}
\usepackage[shortcuts, acronym, hyperfirst = false]{glossaries}
\glsdisablehyper
\usepackage{floatrow}
\newfloatcommand{capbtabbox}{table}[][\FBwidth]
\usepackage{amssymb}
\usepackage[frozencache,cachedir=.]{minted}
\usepackage{algorithm}

\title{Explaining Caption-Image Interactions in CLIP Models \\ with Second-Order Attributions}

\author{\name Lucas Moeller\footnotemark[1] \email lucas.moeller@ims.uni-stuttgart.de \\
      \name Pascal Tilli\thanks{These authors contributed equally.} \email pascal.tilli@ims.uni-stuttgart.de \\
      \name Ngoc Thang Vu \email vu@ims.uni-stuttgart.de \\
      \name Sebastian Pado \email pado@ims.uni-stuttgart.de \\
      \addr University of Stuttgart \\
    }

\def\month{07}  % Insert correct month for camera-ready version
\def\year{2025} % Insert correct year for camera-ready version
\def\openreview{\url{https://openreview.net/forum?id=HUUL19U7HP}} % Insert correct link to OpenReview for camera-ready version

\begin{document}
\newacronym{nlp}{NLP}{Natural Language Processing}
\newacronym{clip}{\textsc{Clip}}{Contrastive Language-Image Pre-Training}
\newacronym{vqa}{VQA}{Visual Question Answering}
\newacronym{coco}{\textsc{Coco}}{Common Objects in Context}
\newacronym{hnc}{\textsc{Hnc}}{Hard Negative Captions}
\newacronym{flickr}{\textsc{Flickr30k}}{}
\newacronym{ig}{IG}{Integrated Gradients}
\newacronym{jxe}{JxE}{Jacobian times Embeddings}
\newacronym{lrp}{\textsc{Lrp}}{Layer-wise relevance propagation}
\newacronym{gradcam}{\textsc{GradCam}}{}
\newacronym{icam}{\textsc{ICam}}{Interaction-\textsc{CAM}}
\newacronym{ilime}{\textsc{ILime}}{Interaction-\textsc{Lime}}
\newacronym{itsm}{\textsc{Itsm}}{}
% \newacronym{volta}{VoLTA}{Vision-Language Transformer with weakly-supervised local-feature Alignment}
\newacronym{cid}{\textsc{Cid}}{}
\newacronym{cii}{\textsc{Cii}}{}
\newacronym{ctd}{\textsc{Ctd}}{}
\newacronym{cti}{\textsc{Cti}}{}
\newacronym{auc}{\textsc{Auc}}{area under the curve}
\newacronym{pg}{\textsc{PG}}{Point Game}
\newacronym{pga}{\textsc{Pga}}{Point Game Accuracy}
\newacronym{pge}{\textsc{Pge}}{Point Game Energy}
\newacronym{sbert}{\textsc{SBert}}{Siamese transformers for
text-text pairs}
\newacronym{rag}{RAG}{Retrieval-Augmented Generation}
\newacronym{pp}{p.p.}{percentage points}

\newcommand{\laion}{\textsc{Laion}~}
\newcommand{\dfn}{\textsc{Dfn}~}
\newcommand{\commonpool}{\textsc{CommonPool}~}
\newcommand{\datacomp}{\textsc{DataComp}~}
\newcommand{\openai}{\textsc{OpenAI}~}
\newcommand{\openclip}{\textsc{OpenClip}~}

\maketitle

\begin{abstract}
  Dual encoder architectures like \textsc{Clip} models map two types of inputs into a shared embedding space and predict similarities between them.
  Despite their wide application, it is, however, not understood \textit{how} these models compare their two inputs.
  Common first-order feature-attribution methods explain importances of individual features and can, thus, only provide limited insights into dual encoders, whose predictions depend on interactions between features.\\
  In this paper, we first derive a second-order method enabling the attribution of predictions by any differentiable dual encoder onto feature-interactions between its inputs. 
  Second, we apply our method to \textsc{Clip} models and show that they learn fine-grained correspondences between parts of captions and regions in images. They match objects across input modes and also account for mismatches.
  This intrinsic visual-linguistic grounding ability, however, varies heavily between object classes, exhibits pronounced out-of-domain effects and we can identify individual errors as well as systematic failure categories.
  % By enabling the analysis of feature-interactions and reaching beyond first-order feature importances, our method contributes to a better understanding of similarity prediction in deep encoder models. 
  Code is publicly available: \url{https://github.com/lucasmllr/exCLIP}
\end{abstract}
\section{Introduction}

\begin{figure*}[tb!]
    \begin{tikzpicture}
        % \node (image) at (0,0) {\includegraphics[width=\textwidth]{plots/slicing_plot.pdf}};
        \node (image) at (0,0) {\includegraphics[width=\textwidth]{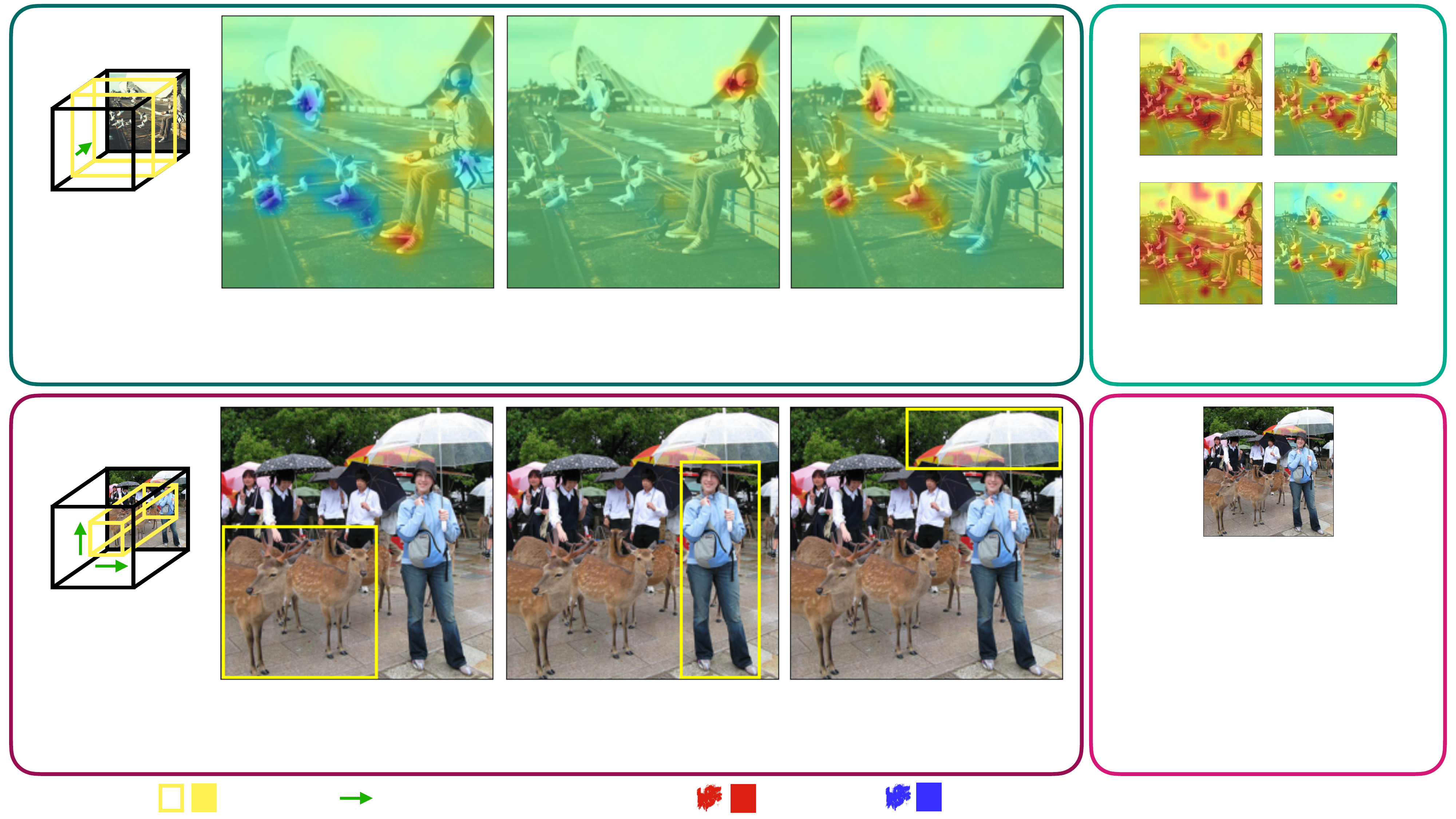}};
        \node at (-0.8, 5.2) {\underline{Our second-order attributions}};
        \node at (-0.8, 4.8) {\underline{\scriptsize Visualization}};
        \node at (-6.9, 4.8) {\underline{\scriptsize Attribution slicing}};
        \node at (6.2, 5.2) {\underline{First-order attributions}};
        \node at (5.4, 2.7) {\scriptsize MaskCLIP};
        \node at (6.9, 2.7) {\scriptsize IG};
        \node at (5.35, 4.4) {\scriptsize CLIPSurgery};
            \node at (6.9, 4.4) {\scriptsize ECLIP};
        % \draw[solid] (4.6, 4.2) -- (4.6, -4.5);
        \node[align=center] at (-4.2, 0.9) {\scriptsize \adjustbox{bgcolor=yellow!100}{\strut A kid} with headphones \\ \scriptsize feeding birds.};
        \node[align=center] at (-1.0, 0.9) {\scriptsize A kid with \adjustbox{bgcolor=yellow!100}{\strut headphones} \\ \scriptsize feeding birds.};
        \node[align=center] at (2.2, 0.9) {\scriptsize A kid with headphones \\ \scriptsize feeding \adjustbox{bgcolor=yellow!100}{\strut birds}.};
        \node[align=center] at (6.1, 0.7) {\scriptsize A kid with headphones \\ \scriptsize feeding birds.};
        \node[align=center] at (-4.2, -3.6) {\scriptsize \adjustbox{bgcolor=red!42}{\strut Deer} \adjustbox{bgcolor=red!0}{\strut next} \adjustbox{bgcolor=red!6}{\strut to} \adjustbox{bgcolor=red!1}{\strut a} \adjustbox{bgcolor=red!1}{\strut woman} \\ \scriptsize \adjustbox{bgcolor=red!4}{\strut with} \adjustbox{bgcolor=red!1}{\strut an} \adjustbox{bgcolor=red!18}{\strut umbrella}.};
        \node[align=center] at (-1.0, -3.6) {\scriptsize \adjustbox{bgcolor=blue!16}{\strut Deer} \adjustbox{bgcolor=red!0}{\strut next} \adjustbox{bgcolor=blue!0}{\strut to} \adjustbox{bgcolor=red!0}{\strut a} \adjustbox{bgcolor=red!29}{\strut woman}  \\ \scriptsize\adjustbox{bgcolor=red!14}{\strut with} \adjustbox{bgcolor=red!3}{\strut an} \adjustbox{bgcolor=red!6}{\strut umbrella}.};
        \node[align=center] at (2.2, -3.6) {\scriptsize \adjustbox{bgcolor=blue!23}{\strut Deer} \adjustbox{bgcolor=blue!1}{\strut next} \adjustbox{bgcolor=blue!0}{\strut to} \adjustbox{bgcolor=blue!0}{\strut a} \adjustbox{bgcolor=blue!3}{\strut woman} \\ \scriptsize\adjustbox{bgcolor=blue!1}{\strut with} \adjustbox{bgcolor=blue!0}{\strut an} \adjustbox{bgcolor=red!63}{\strut umbrella}.};
        \node[align=center] at (6.1, -2.3) {\scriptsize IG \\ \scriptsize \adjustbox{bgcolor=red!42}{\strut Deer} \adjustbox{bgcolor=red!0}{\strut next} \adjustbox{bgcolor=red!6}{\strut to} \adjustbox{bgcolor=red!1}{\strut a} \adjustbox{bgcolor=red!1}{\strut woman} \\ \scriptsize \adjustbox{bgcolor=red!4}{\strut with} \adjustbox{bgcolor=red!1}{\strut an} \adjustbox{bgcolor=red!18}{\strut umbrella}};
        \node[align=center] at (6.1, -3.4) {\scriptsize ECLIP \\ \scriptsize \adjustbox{bgcolor=red!41}{\strut Deer} \adjustbox{bgcolor=red!1}{\strut next} \adjustbox{bgcolor=red!0}{\strut to} \adjustbox{bgcolor=red!0}{\strut a} \adjustbox{bgcolor=red!18}{\strut woman} \\ \scriptsize \adjustbox{bgcolor=red!0}{\strut with} \adjustbox{bgcolor=red!0}{\strut an} \adjustbox{bgcolor=red!34}{\strut umbrella}.};
        \node[align=center] at (-7.2, 2.3) {\scriptsize \(W\)};
        \node[align=center] at (-7.5, 3.8) {\scriptsize \(S\)};
        \node[align=center] at (-7.9, 3.0) {\scriptsize \(H\)};
        \node[align=center] at (-6.5, 4.0) {\scriptsize image};
        \node[align=center, rotate=40] at (-6.3, 2.5) {\scriptsize caption};
        \node[align=center] at (-7.2, -2.2) {\scriptsize \(W\)};
        \node[align=center] at (-7.5, -0.7) {\scriptsize \(S\)};
        \node[align=center] at (-7.9, -1.5) {\scriptsize \(H\)};
        \node[align=center] at (-6.5, -0.5) {\scriptsize image};
        \node[align=center, rotate=40] at (-6.3, -2.0) {\scriptsize caption};
        \node[align=center] at (-7.0, 1.0) {\scriptsize span \\ \scriptsize selection};
        \draw[solid] (-7.0, 1.5) -- (-6.5, 2.65);
        \draw[solid] (-6.5, 1.2) -- (-5.6, 1.1);
        \node[align=center] at (-6.7, -3.5) {\scriptsize bounding-box \\ \scriptsize selection};
        \draw[solid] (-6.5, -3.0) -- (-7.0, -1.65);
        \draw[solid] (-6.5, -3.0) -- (-5.5, -2.8);        
        \node[align=center, rotate=90] at (-8.5, 2.5) {\underline{\scriptsize Image projection}};
        \node[align=center, rotate=90] at (-8.5, -2.0) {\underline{\scriptsize Caption projection}};
        \node[align=center] at (-7.2, -4.4) {\scriptsize Slicing:};
        \node[align=center] at (-5.1, -4.4) {\scriptsize Selection};
        \node[align=center] at (-3.2, -4.4) {\scriptsize Summation};
        \node[align=center] at (-1.3, -4.4) {\scriptsize Heatmaps:};
        \node[align=center] at (1.0, -4.4) {\scriptsize positive};
        \node[align=center] at (3.1, -4.4) {\scriptsize negative};
    \end{tikzpicture}
    \vspace{-0.5cm}
    \caption{(Left column) \textbf{Our second-order attributions can point out \textit{interactions} between arbitrary spans in captions and regions in images}. We can visualize them by slicing (yellow selection) our 3d attribution tensor with image dimensions (\(H,W\)) and caption dimension \(S\) (details in Section \ref{sec:method}). A selection can be projected onto the image (top-left) or the caption (bottom-left) by summation (green arrows). Heatmaps for these projected attributions are in shades of red/blue for positive/negative values. (Right column) In contrast, first-order attributions can only attribute the overall similarity between captions and images onto either the image (top-right) \textit{or} the caption (bottom-right). They \textit{cannot} assess underlying interactions.}
    \label{fig:fist_vs_second_order}
\end{figure*}

Dual encoder models use independent modules to represent two
types of inputs in a common embedding space and are optimized to predict a scalar similarity measure for them. 
The training objective is typically a triplet or contrastive loss \citep{Sohn, Oord}.
Popular examples include \gls{sbert} \citep{sbert} and \gls{clip} models \citep{clip, align} for text-image pairs.
The learned representations have proven to be highly informative for downstream applications, such as image classification \citep{clip_class}, visual question answering \citep{vqa, subg}, image captioning and visual entailment \citep{clip_downstream}, as well as text or image generation \citep{pali, coca, robin}.
In (multi-modal) information retrieval, dual encoders enable efficient semantic search \citep{image-retrieval, llm-ir, splade, ance, gpu-search}, serving e.g. \gls{rag} \cite{rag_survey}
% Their independent processing of the two inputs allows for the pre-computation and storage of representations in vector-databases enabling sub-linear search times via approximate nearest neighbor algorithms.

Despite the wide-spread application of dual encoder models, an open question remains \textit{how} these models compare their two inputs. 
Common first-order attribution methods like Shapley values \citep{shap} or \gls{ig} \citep{ig} can only provide limited insights into dual encoders because they attribute to \textit{individual} features \citep{vis_exp_sim, exp_sim_learners, inthessian, shapleytaylor}. 
However, similarity fundamentally depends on comparisons and, therefore, on \textit{interactions} of features \citep{tversky, lin}.
In dual encoders this manifests in the final cosine-similarity of the two embeddings, resulting in all terms contributing to the output similarity score to contain multiplicative dependencies between the two inputs. 
In such multiplicative terms, a change in one involved feature affects the contribution of others; hence, these features interact.\\
Only few works have studied feature interaction in symmetric Siamese encoders \citep{bilrp, emnlp, eacl, bilrp_naacl} and they have remained almost entirely unstudied in non-symmetric dual encoders like \gls{clip} \citep{ilime}.\\
In this work, we address this research gap and aim at a means to analyze which aspects in two given inputs dual encoders compare in order to predict a similarity for them. 
Our contributions are the following: \\
(1) Motivated by the theory behind \gls{ig} (cf. Appendix \ref{apx:ig}), we derive a general second-order feature attribution method that can explain \textit{interactions} between inputs of
any differentiable dual encoder model. The method does not rely on any
modification of the trained model, nor on additional optimization.
% Required changes to the original code are minimal and easily transferable to different architectures (implementation details in Appendix \ref{apx:implementation}).
% We will make our code available.\\
(2) We apply our method to a range of \gls{clip} models and demonstrate that they can capture fine-grained interactions between corresponding parts of captions and regions in images.
They identify matching objects across the input modes and also penalize mismatches.
Using image-captioning datasets with object bounding-box annotations, we evaluate the extent and limitations of this \textit{intrinsic visual-linguistic grounding ability} in a wide range of \gls{clip} models.
% We find large variation for different object classes and pronounced out-of-domain effects.
% An error analysis reveals typical failure categories. 

Figure \ref{fig:fist_vs_second_order} illustrates our interaction attributions showing how they can point out corresponding parts of captions and regions in images.
In contrast, first-order alternatives cannot access these interactions but only provide insights into aspects important to the \textit{overall} similarity between a text and image input.

% \begin{figure*}
%     \centering
%     \begin{subfigure}[t]{0.49\textwidth}
%         \centering
%         \includegraphics[width=\linewidth]{plots/teaser-figure/image-proj-tmlr.pdf}
%         \caption{Projecting the yellow and red attribution selections on the image by summing over the corresponding sequence tokens.}
%     \end{subfigure}
%     ~
%     \begin{subfigure}[t]{0.49\textwidth}
%         \centering
%         \includegraphics[width=\linewidth]{plots/teaser-figure/cpt-proj-tmlr.pdf}
%         \caption{Projecting the yellow and red attribution selections on the caption by summing over the corresponding patch tokens.}
%     \end{subfigure}
    
%     \caption{
%         Our second-order attributions are illustrated as a cube.
%         (a) slices and aggregates attributions along the sequence axis, which are projected on the image as a heatmap.
%         (b) slices and aggregates attributions from selected bounding boxes, which are projected on the caption, and the largest scores are highlighted in color.
%     }
% \end{figure*}

\section{Related work}

\paragraph{Metric learning}
refers to the task of producing embeddings reflecting the similarity between inputs \citep{dml_survey}.
Applications include face identification \citep{that_you, metric_person} and image retrieval \citep{flickr, soml}.
Siamese networks with cosine similarity of embeddings were early candidates \citep{siamese_metric}.
The triplet-loss \citep{triplet} involving negative examples has been proposed as an improvement but requires sampling strategies for the large number of possible triplets \citep{karsten}.
\citet{soft_triplet} have shown that the triplet-loss can be relaxed to a softmax variant.
\citet{Sohn} and \citet{Oord} have proposed the batch contrastive objective which has been applied in both unsupervised \citep{unsup_contrastive} and supervised representation learning \citep{sup_contrastive}.
It has led to highly generalizable semantic text \citep{sbert} and image embeddings \citep{momentum} and ultimately to the \gls{clip} training paradigm \cite{clip}.

\paragraph{Vision-language models} process both visual and linguistic inputs. 
\citet{medclip} were the first to train a dual-encoder architecture with a contrastive objective on image-text data in the medical domain.
\citet{clip} and \cite{align} have applied this principle to web-scale image captions and alt-text data.
In the following, the basic inter-modal contrastive loss has been extended by intra-modal loss terms \citep{cyclip, uniclip, tcl}, self-supervision \citep{slip}, non-contrastive objectives
\citep{xclip}, incorporating classification labels \citep{unicl}, textual augmentation \citep{laclip}, a unified multi-modal encoder architecture \citep{limoe} and retrieval augmentation \citep{ra-clip}. 
Next to more advanced training objectives, other works have identified the training data distribution to be crucial for performance: \citet{datacomp} have proposed the DataComp benchmark focusing on dataset curation while fixing model architecture and training procedure, \citet{metaclip} have balanced metadata distributions and \citet{dfn} have introduced data filtering networks for the purpose.
The strictly separated dual-encoder architecture has been extended to include cross-encoder dependencies \citep{blip, volta}, and multi-modal encoders have been combined with generative decoders \citep{pali, unifiedio, albef, fromage, flamingo, coca}.

\paragraph{Local feature attribution methods} aim at explaining a given prediction by assigning contributions to individual input features \citep{murdoch, doshi, lipton, atanasova}.
First-order gradients can approximate a prediction's sensitivity to such features
\citep{Li} and gradient\(\times\)input saliencies can approximate feature importance \citep{simonyan}.
In transformer architectures, attention weights have been analyzed \citep{Abnar}, but were subsequently contested as explanation because they are only one component of the model \citep{jain,wiegreffe,bastings}.
\gls{lrp} defines layer-specific rules to back-propagate attributions to individual features \citep{lrp_overview, lrp}.
In contrast, Shapley values \citep{shap} and \gls{ig} \citep{ig} treat models holistically and can provide a form of theoretical guaranty for correctness.
This has recently been challenged by \citet{pnas} who proved fundamental limitations of attribution methods.
A widely used attribution method in the vision domain is \textsc{GradCam} \citep{gradcam}, which \citet{chefer} and \citet{legrad} extended to transformer architectures.\\
Assigning importances to individual features, first-order attribution methods cannot capture dependencies on feature interactions. 
\citet{tsang18} have proposed to detect such interactions from weight matrices in feed-forward neural networks, \cite{bayesia-interaction} investigated them in Bayesian networks.
The Shapley value has been extended to the Shapley (Taylor) Interaction Index \citep{sii, shapleytaylor, bielefeld} and \citet{inthessian} have generalized \gls{ig} to integrated Hessians.
\citet{why_match} and \citet{vis_exp_sim} have assessed interactions underlying similarity predictions in Siamese image encoders.
\citet{bilrp} extended \gls{lrp} for this class of models \citep{bilrp_naacl}, and our prior work extended \gls{ig} to Siamese language encoders \citep{emnlp, eacl}. 
In this work, we further generalize this method to multi-modal dual encoders.

\paragraph{CLIP explainability.}
Several works have previously pursued the goal of better understanding \gls{clip} models and contrastive image encoders.
\cite{clip-bottleneck} and \cite{clip-qda} have proposed information bottleneck approaches. 
\cite{sparse_concepts} identified interpretable sparse concepts in the shared embedding space.
\cite{ecor} predicted human-understandable rationales for images. 
\cite{clip-negation} localized where the text encoder processes negation.
\cite{clip-wordnet} created saliency maps for \textsc{WordNet} concepts.
\cite{gscorecam} proposed an improved CAM variant and analyzed which objects the model looks at.
\cite{disentangle_clip} were interested in the entanglement of image representations.
\citet{clip_disection} identified the roles of individual attention heads in \gls{clip}'s image encoder and later investigated second-order effects of neurons \citep{gandelsman25}. 
\citet{clip_concepts} analyzed whether \gls{clip} models adequately handle compositional concepts.
\citet{clip_differences} investigated the model's reasoning ability about differences in images and \citet{clip_robust} examined safety objectives. 
The unseen performances of \gls{clip} have motivated a number of authors to identify the reasons behind its ostensible generalization ability and robustness towards domain shifts \citep{xue, nguyen, fang, tu, mayilvahanan2024, clip_domain}.
% We discuss these works later, in the light of our own results.
\citet{gradeclip} explored a wide range of first-order methods to attribute similarity scores to images and captions independently and \citet{clip_surgery} proposed the \textsc{CLIPSurgery} method. 
\citet{tip} and \cite{taylorCAM} independently introduced a second-order variant of \textsc{GradCam} that can assess feature interactions. 
It can be applied to \gls{clip}; in Appendix \ref{apx:gradcam}, we show that it is a special case of our method.
Most closely related to our work, \textsc{interactionLime} \citep{ilime} pioneered the attribution of interactions between captions and images in \gls{clip} models.
However, relying on a local bilinear approximation of \gls{clip}, it does not explain the original model and requires additional optimization as well as hyper-parameter tuning (cf. Appendix \ref{sec:ilime}).
Last, \gls{itsm} by \citet{itsm} and the method by \citet{paired_img_sim} are forward-facing saliency methods that compute importance values through pair-wise embedding multiplication.
We compare these approaches against ours in Section \ref{sec:attr_eval}.

\paragraph{Visual-linguistic grounding}\!refers to the identification of
fine-grained relations between text phrases and corresponding image
parts \citep{scene-grounding}.
Specialized models predict regions over
images for a corresponding input phrase \citep{zeroground, cross-ground}.
This objective has been combined with contrastive
caption matching \citep{glip, align2ground}, and caption generation
\citep{unitab}. 
The \textsc{VoLTA} model internally matches latent image-region and
text-span representations \citep{volta}.  
In multi-modal text generative models, grounding has been included as an
additional pretraining task \citep{unicode-vl, vl-bert, uniter} and can be unlocked with visual prompt
learning \citep{michi}.
At the intersect of grounding and explainability, \citet{gen_vis_expl} have generated textual explanations for vision models and have grounded them to input images \citep{ground_vis_expl, multimodal_expl}.
In this paper, we do not optimize models to explicitly ground predictions, but
aim at analyzing to which extent purely contrastively trained dual encoders acquire this ability intrinsically.

\section{Method} \label{sec:method}
We first derive general second-order attributions for dual encoder predictions enabling the assessment of feature-interactions between their two inputs. In the following, we then describe their realization in transformer models, specifically \gls{clip}.
\paragraph{Derivation of second-order attributions.}
We begin from the definition of a dual encoder \(f\),
\begin{equation} \label{eq:model}
    s = f(\mathbf{a}, \mathbf{b}) = \mathbf{g}(\mathbf{a})^\top \mathbf{h}(\mathbf{b}),
\end{equation} 

with two vector-valued encoders \(\mathbf{g}\) and \(\mathbf{h}\), respective inputs \(\mathbf{a}\) and \(\mathbf{b}\) and a scalar output \(s\).
% For our purpose, \(\mathbf{g}\) will be an image encoder with an image input \(\mathbf{a}\) and \(\mathbf{h}\) will be a text encoder with a text representation \(\mathbf{b}\) as input.
We also define two \textit{reference} inputs \(\mathbf{r}_a\) and \(\mathbf{r}_b\), whose role will be discussed later.
With these definitions, we can write the following expression,
\begin{equation} \label{eq:ansatz}
    f(\mathbf{a}, \mathbf{b}) - f(\mathbf{r}_a, \mathbf{b}) - f(\mathbf{a}, \mathbf{r}_b) + f(\mathbf{r}_a, \mathbf{r}_b),
\end{equation}
which will serve as a rigorous starting-point of our derivation. 
In the following, we first proceed by showing the equality of this initial starting-point to Eq. \ref{eq:attr_mat}. 
We then reduce this equality to our final result in Eq. \ref{eq:appr_attr} using the approximations discussed below.
At this point, we are also discussing an intuitive interpretation of the final result.\\
As a first step, we see \(f\) as an anti-derivative and reformulate the above expression into an integral over its derivative:
\begin{equation} \label{eq:integral}
\begin{split}
    \big[ f(\mathbf{a}, \mathbf{b}) - f(\mathbf{r}_a, \mathbf{b}) \big] &-  \big[ f(\mathbf{a}, \mathbf{r}_b) - f(\mathbf{r}_a, \mathbf{r}_b) \big] \\
    = \int_{\mathbf{r}_b}^\mathbf{b} \, \frac{\partial}{\partial \mathbf{y}_j} \, \big[ f(\mathbf{a}, \mathbf{y}) - f(\mathbf{r}_a, \mathbf{y}) \big] \, d\mathbf{y}_j
    & = \int_{\mathbf{r}_b}^\mathbf{b}\! \int_{\mathbf{r}_a}^\mathbf{a} \frac{\partial^2}{\partial \mathbf{y}_j \partial \mathbf{x}_i} \, f \left(\mathbf{x}, \mathbf{y}\right) \,d\mathbf{x}_i \,d\mathbf{y}_j
\end{split}
\end{equation}
Here, \(\mathbf{x}\) and \(\mathbf{y}\) are integration variables for the two inputs.
This step can be seen as the second-order equivalent to Equation \ref{eq:int_grad_diff} in the theory behind \gls{ig} (cf. Appendix \ref{apx:ig}).
We use component-wise notation with indices \(i\) and \(j\) for the input dimensions and omit sums over double indices for clarity.
We then plug in the model definition from Equation \ref{eq:model},
\begin{equation} \int_{\mathbf{r}_a}^\mathbf{a}\!\int_{\mathbf{r}_b}^\mathbf{b} \frac{\partial^2}{\partial \mathbf{x}_i \partial \mathbf{y}_j} \, \mathbf{g}_k (\mathbf{x})\, \mathbf{h}_k(\mathbf{y}) \,d\mathbf{x}_i \,d\mathbf{y}_j,
\end{equation}
again using component-wise notation for the dot-product with \(k\) indexing the dimension of the shared embedding space.
Since neither embedding depends on the other integration variable, we can separate both integrals and derivatives applying a product rule:
\begin{equation} \label{eq:sep_int}
 \int_{\mathbf{r}_a}^\mathbf{a} \frac{\partial \mathbf{g}_k (\mathbf{x})}{\partial \mathbf{x}_i}\,d\mathbf{x}_i \int_{\mathbf{r}_b}^\mathbf{b} \frac{\partial \mathbf{h}_k(\mathbf{y})}{\partial \mathbf{y}_j} \,d\mathbf{y}_j
\end{equation}
% 
% This step makes explicit use of the strict independence of the two encoders.
% Cross-encoder architectures would introduce dependencies between them.
Both terms are line integrals from the references to the actual inputs in the respective input representation spaces;
\(\partial \mathbf{g}_k (\mathbf{x}) / \partial \mathbf{x}_i\) and \(\partial \mathbf{h}_k (\mathbf{y}) / \partial \mathbf{y}_j\) are the Jacobians of the two encoders.
To proceed with these integrals, we define integration paths and substitute. 
We follow \citet{ig}, and use the straight lines between both references and inputs,
\begin{align} \label{eq:lines}
    \mathbf{x}(\alpha) &= \mathbf{r}_a + \alpha (\mathbf{a} - \mathbf{r}_a),\\
    \mathbf{y}(\beta) &= \mathbf{r}_b + \beta (\mathbf{b} - \mathbf{r}_b),
\end{align}
parameterized by \(\alpha\) and \(\beta\), respectively.
For the integral over encoder \(\mathbf{g}\) substituting the path \(\mathbf{x}(\alpha)\) yields an integral over the scalar integration variable \(\alpha\):
\begin{equation} \label{eq:substitution}
    \int_{0}^{1} \frac{\partial \mathbf{g}_k \left(\mathbf{x(\alpha})\right)}{\partial \mathbf{x}_i}\,\frac{\partial \mathbf{x}_i(\alpha)}{\partial \alpha} \, d\alpha \\
    =\, (\mathbf{a} - \mathbf{r}_a)_i \int_{0}^{1} \frac{\partial \mathbf{g}_k \left(\mathbf{x(\alpha})\right)}{\partial \mathbf{x}_i} \, d\alpha
\end{equation}
Since \(\partial \mathbf{x}(\alpha) / \partial \alpha = (\mathbf{a} - \mathbf{r}_a)\) is a constant w.r.t \(\alpha\), we can pull it out of the integral. 
We then define the \textit{integrated Jacobian} for the encoder \(\mathbf{g}\),
\begin{equation} \label{eq:int_jac}
    \mathbf{J}_{ki}^g \coloneqq \int_{0}^{1} \frac{\partial \mathbf{g}_k \left(\mathbf{x(\alpha})\right)}{\partial \mathbf{x}_i} \, d\alpha
    \approx \frac{1}{N} \, \sum_{n=1}^N \, \frac{\partial \mathbf{g}_k(\mathbf{x}(\alpha_n))}{\partial \mathbf{x}_i},
\end{equation}
as the analogon to integrated gradients for vector-valued models. 
The integral over encoder \(\mathbf{h}\) can be processed in the same way by substituting \(\mathbf{y}(\beta)\) to obtain \(\mathbf{J}_{kj}^h\). 
In practice, these integrals are calculated numerically by sums over \(N\) steps, with \(\alpha_n = n/N\). This introduces an approximation error which must, however, converge to zero for large \(N\) by definition of the Riemann integral.
We plug the results from Equation \ref{eq:substitution} and the definitions of the \textit{integrated Jacobians} into Equation \ref{eq:sep_int}:
\begin{equation} \label{eq:attr_mat}
    (\mathbf{a} - \mathbf{r}_a)_i \,\mathbf{J}_{ik}^g \mathbf{J}_{kj}^h \, (\mathbf{b} - \mathbf{r}_b)_j \eqcolon \sum_{ij} \mathbf{A}_{ij}
\end{equation}

After computing the sum over the output embedding dimension \(k\), this yields interaction terms, each involving a feature pair \((i, j)\) with feature \(i\) from input \(\mathbf{a}\) and feature \(j\) from input \(\mathbf{b}\).
We can write the values of these terms for all feature pairs into a matrix with index \(i\) on one side and \(j\) on the other, which we refer to as the \textit{attribution matrix} \(\mathbf{A}_{ij}\).
In the last step, we write out the omitted sum over \(i\) and \(j\) explicitly.
% This final result is the product between the integrated Jacobians of the two encoders, thus, reducing the initial second-order derivative from Eq. \ref{eq:integral} to a product of two first-order derivatives.\\
Note that except for the numerical integration, the equality to Equation \ref{eq:ansatz} still holds. 
Hence, the sum over all feature pair attributions in \(\mathbf{A}\) is an exact reformulation of our starting-point.\\
At this point, we return to the \textit{references}, \(\mathbf{r}_a\) and \(\mathbf{r}_b\), defined above. 
We require them to be approximately dissimilar to any other input, e.g. a black image or a caption consisting of padding tokens for respective encoders.
If this is the case all three terms involving \(\mathbf{r}_a\) and \(\mathbf{r}_b\) in Equation \ref{eq:ansatz} approximately vanish, i.e.
\(
    f(\mathbf{r}_a, \mathbf{b}) \approx 0, \, f(\mathbf{a}, \mathbf{r}_b) \approx 0, \, f(\mathbf{r}_a, \mathbf{r}_b) \approx 0.
\) 
This reduces the equality between Equations \ref{eq:ansatz} and \ref{eq:attr_mat} to our final result:
\begin{equation} \label{eq:appr_attr}
    f(\mathbf{a}, \mathbf{b}) \approx \sum_{ij} \mathbf{A}_{ij}.
\end{equation}
Intuitively, this provides an approximate decomposition of the model prediction \(s\!=\!f(\mathbf{a}, \mathbf{b})\) into additive contributions of feature-pair interactions between the two inputs.
% It can be thought of as the natural extension of integrated gradients \citep{ig} to dual encoder models.
Throughout this paper, we evaluate the attribution matrix \(\mathbf{A}\). 
% Below, we describe which form it takes in \gls{clip} models and how we proceed to visualize it.

\paragraph{Interaction attributions in transformer models.}
In the derivation above, we treat image and text representations as vectors.
In transformer-based encoders, text inputs are represented as \(S \!\times\! D_g\) dimensional tensors, where \(S\) is the length of the token sequence. 
Image representations are of shape \(H \!\times\! W \!\times\! D_h\), with \(H\) and \(W\) being height and width of the image representation; in vision-transformers both equal the number of patches \(P\).
\(D_g\) and \(D_h\) are the encoders' embedding dimensionalities.
Our pair-wise image-text interaction attributions thus have the dimensions \(H \!\times\! W \!\times\! D_h \!\times\! S \!\times\! D_g\), which quickly becomes intractably large. 
Fortunately, the sum over dimensions in Equation \ref{eq:appr_attr} enables the additive combination of attributions in \(\mathbf{A}\).
We sum over the embedding dimensions of both encoders \(D_g\) and \(D_h\) and obtain an \(H \!\times\! W \!\times\! S\) dimensional attribution tensor, which estimates for \textit{each pair of a text token and an image patch} how much their interaction contributes to the overall prediction. 
These attributions are still three-dimensional and thus not straightforward to visualize.
However, again we can use their additivity, slice the 3d attribution tensor along text or image dimensions and project onto the remaining dimensions by summation.
This projection is demonstrated in Figure \ref{fig:fist_vs_second_order} schematically and with examples, both for a selection over a token range in the caption (top) and a selection over a bounding-box in the image (bottom).
% Importantly, all three examples per caption-image pair come from the same 3d attribution tensor.
Albeit \gls{clip} models are typically trained to match images against captions, we can also compute intra-modal attributions for image-image or text-text pairs by applying the same encoder to both inputs.
Appendix \ref{apx:intra-modal} discusses this in more detail.
Figure \ref{fig:intra-modal} shows two examples.
\begin{table}[tb!]
    \begin{minipage}{0.54\textwidth}
        \centering
        \small
        \begin{tabular}{p{7.0cm}}
             A \adjustbox{bgcolor=yellow!100}{\strut hot dog} sitting on a table covered in confetti. \vspace{.1cm}\\
             \adjustbox{bgcolor=blue!0}{\strut Surrounded} \adjustbox{bgcolor=red!0}{\strut by} \adjustbox{bgcolor=red!0}{\strut glitter,} \adjustbox{bgcolor=red!2}{\strut there} \adjustbox{bgcolor=blue!0}{\strut is} \adjustbox{bgcolor=red!0}{\strut a} \adjustbox{bgcolor=red!63}{\strut sausage} \adjustbox{bgcolor=red!9}{\strut in} \adjustbox{bgcolor=red!2}{\strut a} \adjustbox{bgcolor=red!7}{\strut bun}. \vspace{.1cm}\\
             \hline \vspace{.1cm}
             A hot dog sitting on a table covered in \adjustbox{bgcolor=yellow!100}{\strut confetti}. \vspace{.1cm}\\
             \adjustbox{bgcolor=red!7}{\strut Surrounded} \adjustbox{bgcolor=blue!0}{\strut by} \adjustbox{bgcolor=red!29}{\strut glitter}, \adjustbox{bgcolor=red!18}{\strut there} \adjustbox{bgcolor=blue!0}{\strut is} \adjustbox{bgcolor=blue!0}{\strut a} \adjustbox{bgcolor=blue!5}{\strut sausage} \adjustbox{bgcolor=blue!2}{\strut in} \adjustbox{bgcolor=blue!0}{\strut a} \adjustbox{bgcolor=blue!8}{\strut bun}.\\
        \end{tabular}
        \vspace{.5cm}
    \end{minipage}
    \begin{minipage}{0.44\textwidth}
        \hspace{.2cm}
        \begin{tikzpicture}
        \node (image) at (0,0) {\includegraphics[width=.9\textwidth]{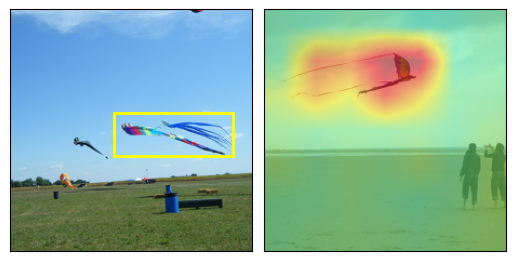}};
        \draw[solid] (-3.5, 1.6) -- (-3.5, -1.6);
        \end{tikzpicture}
        \vspace{.05cm}
    \end{minipage}
    \captionof{figure}{(Left) Intra-modal text-text attributions between top and bottom captions (top: selections in yellow, bottom: corresponding attributions in red/blue for positive/negative). (Right) Intra-modal image-image attributions between left and right image (left: bounding-box selection in yellow, right: heatmaps as above). More examples can be found in Figure \ref{fig:more_img_img}.}
    \label{fig:intra-modal}
\end{table}
\section{Experiments} \label{sec:setting} 
In our experiments, we apply our feature-interaction attributions to \gls{clip} models.
We focus on evaluating the interactions between mentioned objects in captions and corresponding regions in images by selecting token-ranges in captions and analyzing their interactions with image patches.
In the first series of experiments, we compare our attributions against baselines (Section \ref{sec:attr_eval}). The second series in Section \ref{sec:model_analysis} then utilizes our method and analyzes \gls{clip} models.

\subsection{Experimental setting}
We base our evaluation on three image-caption datasets that additionally contain object bounding-box annotations in images, Microsoft's \gls{coco} \citep{coco}, the \gls{flickr} collection \citep{flickr30k} with entity annotations \citep{flickr30k_entities}, and the \gls{hnc} dataset by \citet{hnc}.
\gls{hnc} generates captions from scene graphs based on templates.
We use a basic \texttt{subject predicate object} template to align with the domain of the other two datasets.
We use \gls{hnc} for evaluation only, on \gls{flickr} we use the test split, and on \gls{coco} we use the validation split for our analysis as the test split does not contain captions\footnote{\hyperlink{https://www.kaggle.com/datasets/shtvkumar/karpathy-splits}{https://www.kaggle.com/datasets/shtvkumar/karpathy-splits}}.

We work with \gls{clip} dual encoders \citep{clip} trained with the standard inter-modal contrastive objective and analyze the original \openai models, as well as
\textsc{MetaClip} \citep{metaclip} and the \openclip reimplementations trained on the \laion \citep{laion}, \dfn \citep{dfn}, \commonpool, and \datacomp \citep{datacomp} datasets\footnote{
CLIP family: \hyperlink{https://github.com/openai/CLIP}{https://github.com/openai/CLIP}, Open family: \hyperlink{https://github.com/mlfoundations/open_clip}{https://github.com/mlfoundations/open\_clip}
}.
If not mentioned otherwise, our experiments are based on the ViT-B-16 architecture.
In addition to the unmodified models, we evaluate variants fine-tuned on the \gls{coco} and \gls{flickr} training splits. 
We run all trainings for five epochs using AdamW \citep{adamw}, starting with an initial learning rate of \num{1e-7} that exponentially increases to \num{1e-5}.
Weight decay is set to \num{1e-4} and the batch size is \num{64} on a single 50GB \textsc{Nvidia} A6000.

\subsection{Attribution evaluation} \label{sec:attr_eval}

In the first series of experiments, we compare our attributions against baselines. 
Figure \ref{fig:fist_vs_second_order} includes a qualitative comparison of our second-order interaction attributions against first-order variants.
A detailed comparison between first-order methods has been
presented by \cite{gradeclip}. We closely follow their evaluation
protocol and extend it to second-order methods.
Unless stated otherwise, we attribute to the second-last hidden representation in the models' image and text encoders and use \(N\!=\!50\) integration steps (cf. Equation \ref{eq:int_jac}), with a black image as the image reference and a padding token sequence as the text reference.
In Appendix \ref{apx:add_exp}, we include additional experiments on the accuracy of our attributions as a function of \(N\), as well as different reference choices.

\paragraph{Baselines.}
We compare our method against four baselines: \gls{icam} \citep{tip} is also gradient-based and can be seen as a special case of our approach as shown in Appendix \ref{apx:gradcam}.
\gls{ilime} is a bilinear extension of \textsc{Lime} for dual-encoder models \citep{ilime}. 
Code is not available, therefore, we reimplement it; details are in Appendix \ref{sec:ilime}.
\gls{itsm} \citep{tip} follows the simple approach of pair-wise multiplication of token and image patch embeddings after applying \gls{clip}'s final projection layer to the individual embeddings. 
Originally, it is applied to output representations and we refer to this variant as \textsc{Itsm-O}.
We also apply it to the same hidden representations that our method attributes to and refer to this variant as \textsc{Itsm-H}.
A qualitative comparison between all methods is included in Figure \ref{fig:qualitative}.
None of the used methods including our own modifies the model architecture, its parameters, embeddings or gradients.

\begin{figure}
\begin{floatrow}
    \ffigbox{
    \includegraphics[width=.9\linewidth]{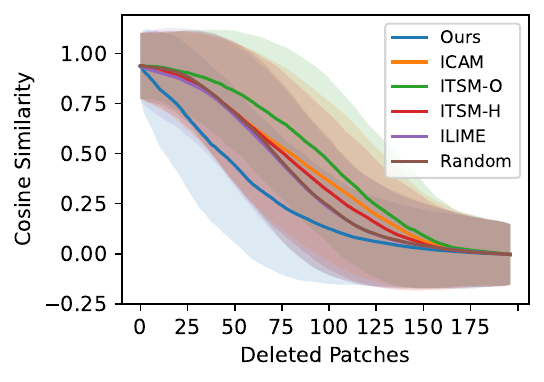}
    }{
    \caption{\label{fig:perturb}
    \textbf{Decline of average similarity scores for iterative image patch deletions according to attributions} for the \laion model fine-tuned on \textsc{Coco}. Uncertainty intervals are standard deviation over the evaluation split.}
    }
    \capbtabbox[0.94\FBwidth]{
    \vspace{0.22cm}
    \resizebox{1.0\linewidth}{!}{
    \begin{tabular}{lrrrrrr}
    \toprule
     & Ours & \textsc{Icam} & \textsc{Itsm-O} & \textsc{Itsm-H} & \textsc{ILime} & Random \\
    \midrule
    \multicolumn{7}{c}{\laion (tuned)} \\
    \midrule
    \gls{cid} $\downarrow$ & \textbf{65.94} & 91.39 & 101.35 & 88.88 & 83.43 & 85.04 \\
    \gls{cii} $\uparrow$   & \textbf{113.35} & 80.21 & 69.01 & 87.53 & 86.48 & 85.18 \\
    \gls{ctd} $\downarrow$ & 4.32 & 7.86 & 6.28 & 7.41 & \textbf{4.05} & 7.52 \\
    \gls{cti} $\uparrow$   & 9.26 & 7.77 & 7.82 & 6.72 & \textbf{9.89} & 7.47 \\
    \midrule
    \multicolumn{7}{c}{\openai} \\
    \midrule
    \gls{cid} $\downarrow$ & \textbf{16.33} & 21.04 & 23.97 & 20.91 & 20.25 & 20.47 \\
    \gls{cii} $\uparrow$   & \textbf{24.43} & 20.47 & 16.24 & 20.60 & 20.58 & 20.47 \\
    \gls{ctd} $\downarrow$ & \textbf{1.06} & 1.35 & 1.15 & 1.30 & 1.07 & 1.31 \\
    \gls{cti} $\uparrow$   & \textbf{1.06} & 0.95 & 0.99 & 0.92 & 1.04 & 0.96 \\
    \bottomrule
    \end{tabular}
    }
    }{
    \caption{\label{tab:perturbs} \textbf{The AUC for CID, CII, CTD, and CTI}, on \gls{coco} for the fine-tuned \laion and the original \openai model. $\downarrow$: lower is better; $\uparrow$: higher is better. Corresponding plots in Fig. \ref{app:fig:conditional_insert_deletion_laion}.}
    }
\end{floatrow}
\end{figure}

\paragraph{Input perturbation.}
Following \citet{tip}, we perform conditional perturbation experiments by iteratively removing or inserting the most attributed features in one input while keeping the other input unmodified.
Figure \ref{fig:perturb} plots the decrease in similarity score for \gls{cid}. 
Our method produces the steepest score decline as a function of the number of patches removed, indicating its ability to identify the most relevant interactions.
Next to \gls{cid}, we also evaluate conditional image patch insertion (\gls{cii}) as well as text token deletion (\gls{ctd}) and insertion (\gls{cti}). 
All plots are shown in Figures \ref{app:fig:conditional_insert_deletion_laion} and \ref{app:fig:conditional_insert_deletion_openai}. 
Table \ref{tab:perturbs} provides a summary and reports the \gls{auc} for the four variants.
With the exception of \gls{ilime} on the text side, our method consistently results in the highest \gls{auc} values for the insertion experiments and the lowest for deletion.
While \gls{ilime} performs well on conditional text attribution, interestingly, its image attributions are not competitive. We discuss this in Appendix \ref{sec:ilime}.
Insertion and deletion experiments have been criticized for producing out-of-domain inputs \citep{hooker}.
Therefore, we also construct in-domain perturbations through  \textit{hard negative captions} and evaluate their effects in Section \ref{sec:model_analysis}.

\thisfloatsetup{subfloatrowsep=none}

\begin{figure*}
    \centering
    \begin{tikzpicture}
      \node (image) at (0,0) {\includegraphics[width=1.0\textwidth]{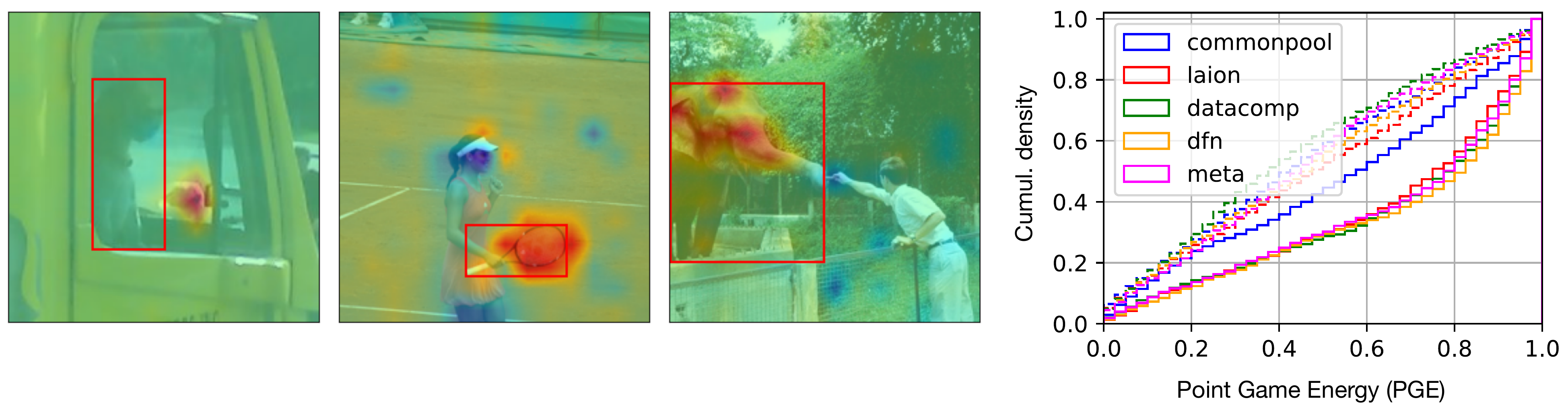}};
      \node[align=center] at (-6.5, -1.8) {\scriptsize A \adjustbox{bgcolor=yellow!100}{\strut dog} sitting inside \\ \scriptsize the cab of a truck. \\ \scriptsize (\gls{pge}=0.1, \gls{pga}=0)};
      \node[align=center] at (-3.0, -1.8) {\scriptsize A woman holding her\\ \scriptsize tennis \adjustbox{bgcolor=yellow!100}{\strut racket}. \\ \scriptsize  (\gls{pge}=0.4, \gls{pga}=1)};
      \node[align=center] at (0.5, -1.8) {\scriptsize A guy feeding an  \\ \scriptsize\adjustbox{bgcolor=yellow!100}{\strut elephant} over a fence. \\ \scriptsize (\gls{pge}=0.9, \gls{pga}=1)};
    \end{tikzpicture}
    \caption{(Left) \textbf{Examples for attributions between selected objects in the caption (yellow) and the image} together with corresponding \gls{coco} bounding-boxes (red), \gls{pge} and \gls{pga} values as described in Section \ref{sec:bbox_attr}. (Right) \textbf{Cumulative PGE distributions} for the \openclip models on \gls{coco} before (dashed) and after (solid) in-domain fine-tuning.}
    \label{fig:pge}
\end{figure*}

\paragraph{Object localization.} \label{sec:bbox_attr}
Following \citet{gradeclip}, we employ the \gls{pg} framework by \citet{pg} to evaluate how well attributions between objects mentioned within captions and images, correspond to human bounding-box annotations. 
In \gls{flickr}, spans in captions that correspond to bounding-boxes are already annotated.
In \gls{hnc}, object classes exactly match sub-strings in captions and 
for \gls{coco}, we identify objects in captions through a dictionary based synonym matching. 
For this experiment, we include all object annotations that correspond to a single instance of its class in the image, and whose bounding-box is larger than one patch. 
This results in 3.5k image-caption pairs from \gls{coco}, 8k pairs from \gls{flickr}, and 500 pairs from \gls{hnc}.
% \paragraph{Localization evalutaion.}
Within the \gls{pg}-framework, \gls{pga} defines the fraction of cases for which the most attributed patch falls within the objects' bounding-box, and \gls{pge} is the fraction of positive attributions within the bounding-box relative to the total attribution \citep{odam, scorecam}.
For \gls{pge}, we compare both full distributions (Figure \ref{fig:pge} (right)) and median values (m\gls{pge}).
Figure \ref{fig:pge} shows examples from different \gls{pge}-ranges and the corresponding cumulative distributions.
Very high or low values, unambiguously indicate good correspondence or clear failure cases, respectively.
Intermediate values often result from attributions extending to contextual elements beyond actual bounding boxes, such as the \emph{tennis court} in the second example.

Table \ref{tab:baselines} compares our method against the baselines.
Full results are in Table \ref{tab:full_baselines}.
\textbf{Our attributions outperform the baselines by large margins.}
Figure \ref{fig:pge_hist_ablation} includes corresponding cumulative \gls{pge}-distributions.
Based on these distributions, we test whether improvements are statistically significant using the framework of stochastic order (\citet{dror}; details in Appendix \ref{sec:stoch_order}).
At the strict criterion of \(p\!<\!0.001\) and \(\epsilon\!=\!0.01\), our method consistently results in significantly better \gls{pge}-statistics.
% This shows that neither a simplified gradient-based approach (\gls{icam}), nor pair-wise embedding multiplication (\gls{itsm}) or the optimization of a local surrogate model (\gls{ilime}) can capture caption-image interactions in \gls{clip} models as well as our method.

\begin{table}[tb!]
    \begin{subtable}[T]{.49\textwidth}
    \resizebox{\linewidth}{!}{
        \begin{tabular}{rlcccc}
            \toprule
             \multicolumn{2}{c}{} & \multicolumn{2}{c}{\gls{coco}} & \multicolumn{2}{c}{\gls{flickr}} \\
            \cmidrule(r){3-4} \cmidrule(r){5-6}
            \textbf{Training} & \textbf{Method} & m\gls{pge} & \gls{pga} & m\gls{pge} & \gls{pga} \\
            \midrule
            \multirow{4}{*}{\openai } & \textsc{Itsm-O} & 18.1 & 21.4 & 19.5 & 23.3 \\
            & \textsc{Itsm-H} & 29.8 & 38.1 & 29.5 & 37.4 \\
            & \textsc{Ilime} & 27.9 & 34.9 & 25.8 & 33.1 \\
            & \textsc{Icam} & 38.6 & 54.6 & 33.5 & 51.4 \\
            & Ours & \textbf{72.3} & \textbf{79.0} & \textbf{64.4} & \textbf{72.1} \\
            \hline
            \multirow{4}{*}{\textsc{Laion} (tuned)} & \textsc{Itsm-O} & 22.8 & 30.3 & 24.5 & 28.7 \\
            & \textsc{Itsm-H} & 30.5 & 34.6 & 28.8 & 36.6 \\
            & \textsc{Ilime} & 28.8 & 37.8 & 25.8 & 34.5 \\
            & \textsc{Icam} & 32.5 & 58.4 & 33.5 & 51.4 \\
            & Ours & \textbf{71.1} & \textbf{83.2} & \textbf{54.3} & \textbf{61.8} \\
            \bottomrule
        \end{tabular}
   }
   \caption{\textbf{\gls*{pg}-based comparison of our attributions against all baselines} described above.}
    \label{tab:baselines}
   \end{subtable}
   \hfill
   \begin{subtable}[T]{.49\textwidth}
        \setlength{\tabcolsep}{4pt}
        \resizebox{\linewidth}{!}{
            \begin{tabular}{rccccccc}        
                \toprule
                \multicolumn{2}{c}{} & \multicolumn{2}{c}{\gls{coco}} & \multicolumn{2}{c}{\gls{hnc}} & \multicolumn{2}{c}{\gls{flickr}} \\
                \cmidrule(r){3-4} \cmidrule(r){5-6} \cmidrule(r){7-8}
                \textbf{Training} & \textbf{Tuning} & m\textsc{Pge} & \textsc{Pga} & m\textsc{Pge} & \textsc{Pga} & m\textsc{Pge} & \textsc{Pga} \\
                \midrule
                \multirow{2}{*}{\openai } & No & 72.3 & 79.0 & 57.0 & 65.0 & 64.4 & 72.1 \\
                & Yes & \textbf{78.0} & \textbf{82.9} & - & - & \textbf{73.4} & \textbf{79.0} \\
                \multirow{2}{*}{Laion} & No & 49.4 & 63.3 & 40.0 & 51.6 & 38.2& 52.0 \\
                & Yes & \textbf{71.1} & \textbf{83.2} & - & - & \textbf{54.6} & \textbf{61.8} \\
                \bottomrule
            \end{tabular}
       }
       \caption{\textbf{Results of the PG-based grounding evaluation} for the  \openai and \laion models. \textit{Tuning} indicates whether a model was fine-tuned on the respective train split of a dataset. Improvements upon fine-tuning are in bold.}
       \label{tab:pg_results}
   \end{subtable}
   \caption{\gls{pga}: Point Game Accuracy, m\gls{pge}: median Point Game Energy. Extensive results for Table \ref{tab:baselines} including additional models are shown in Table \ref{tab:full_baselines}. Full results of Table \ref{tab:pg_results} can be found in Tables \ref{tab:openai_bbox_attr} and \ref{tab:openclip_bbox_attr}.}
\end{table}

\subsection{Model analysis} \label{sec:model_analysis}
We now turn to applying our method to gain insights into how \gls{clip} models match images and captions.

\paragraph{Intrinsic grounding ability.}
Many of the tested models achieve good performances on the object localization task.
On \gls{coco}, the off-the-shelf \openai (fine-tuned \laion\!) points to the correct objects in images in \(79.0\%\) (\(83.2\%\)) of the cases (\gls{pga}) and their high \gls{pge} values show that overall, attributions are distributed to the correct image regions.
Table \ref{tab:pg_results} as well as Table \ref{tab:openai_bbox_attr} and \ref{tab:openclip_bbox_attr} include the results for all models and datasets.
We emphasize that all models including the fine-tuned ones, have only been trained on contrastive caption-image matching.
Therefore, the strong intrinsic grounding abilities that we observe here show that \textbf{the coarse contrastive objective can induce fine-grained correspondence between caption parts and image regions in CLIP models}.
However, we also observe large differences between the original models and the fine-tuned variants, especially in the \openclip models.

\paragraph{Out-of-domain effects.}
The off-the-shelf models were trained on large web-based captioning datasets but have (presumably) not been exposed to the \gls{flickr} and \gls{coco} train splits.
To assess domain effects of their grounding abilities, we compare the versions fine-tuned on respective train splits to the original models in Table \ref{tab:pg_results} and the Appendix.
Figure\ \ref{fig:pge} also plots cumulative \gls{pge} distributions for both.
While the unmodified \openai model already demonstrates strong grounding abilities on \gls{coco} and \gls{flickr}, the off-the-shelf \openclip counterparts perform notably worse.
Upon in-domain fine-tuning the \openclip models improve by an average of \(21.7\!\pm\!8.3\) (\(14.4\!\pm\!4.3\)) \gls{pp} in median \gls{pge} and by \(18.7\!\pm\!6.2\) (\(9.18\!\pm\!4.4\)) \gls{pp} in \gls{pga} on \gls{coco} (\gls{flickr}).
All changes are significant at \(p\!<\!0.001\) and \(\epsilon\!=\!0.01\).
These large improvements indicate that \textbf{this fine-grained connection between captions and images, however, struggles to generalize beyond the training domain}.

\paragraph{Class-wise evaluation.} 
\begin{figure*}
    \centering
    \begin{tikzpicture}
      \node (image) at (0,0) {\includegraphics[width=1.0\textwidth]{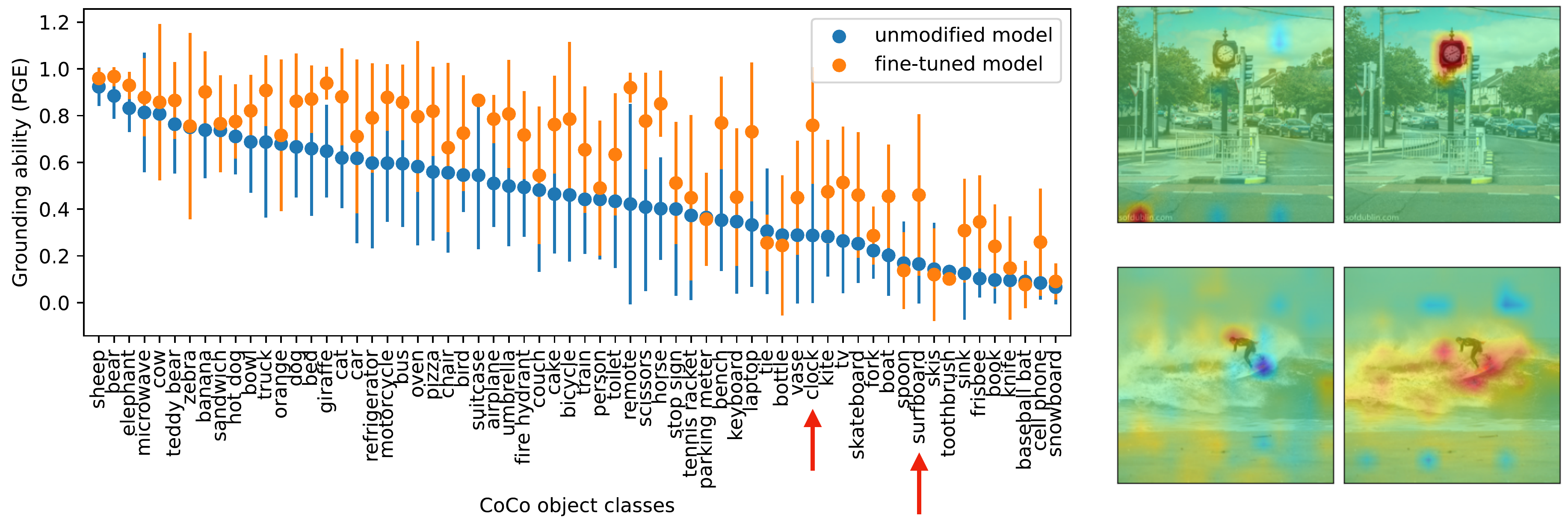}};
      \node[align=center] at (5.8, 0.2) {\scriptsize A \adjustbox{bgcolor=yellow!100}{\strut clock} in the intersect of two streets.};
      \node[align=center] at (5.4, -2.5) {\scriptsize A surfer riding a wave on a yellow \adjustbox{bgcolor=yellow!100}{\strut surfboard}.};
      \node[align=center] at (4.6, 2.85) {\scriptsize \underline{unmodified}};
      \node[align=center] at (7.0, 2.85) {\scriptsize \underline{fine-tuned}};
    \end{tikzpicture}
    \caption{(Left) \textbf{Class-wise average PGE before and after in-domain fine-tuning} in the \openclip \laion model on \gls{coco}. Error bars are standard deviations over all class instances. (Right) Two explicit examples of how the model's grounding ability changes upon tuning. The corresponding classes are emphasized with red arrows on the left.}
    \label{fig:pge_classes}
\end{figure*}
The generalization issue of the models' grounding ability becomes apparent in the examples shown in Figure \ref{fig:pge_classes}.
The off-the-shelf model fails to identify the \emph{clock} and even assigns a negative attribution to the \emph{surfboard}, whereas the fine-tuned version clearly identifies both. 
To examine the models' understanding of individual visual-linguistic concepts in more detail, we break the above analysis down to individual classes.
The right side of Figure \ref{fig:pge_classes} shows average \gls{pge}-values and their standard deviations for \gls{coco} classes in the \openclip \laion model.
The classes are ordered from left to right based on their average grounding ability in the unmodified model (blue).  
The model effectively identifies the leftmost classes, while grounding is notably weaker for the rightmost. 
Upon fine-tuning (orange), most classes show clear improvements.
By means of the standardized mean difference between the two \gls{pge} values, we observe the largest improvements for the classes \emph{horse}, \emph{bench}, \emph{giraffe}, \emph{airplane} and \emph{clock}.
This shows that \textbf{contrastive fine-tuning can sharpen the visual-linguistic conception of individual object classes}.
In Appendix \ref{apx:add_results} (Figure \ref{fig:add_class_gamma}), we replicate this experiment for \dfn and \commonpool yielding similar results.

\begin{figure}[tb!]
    \centering
    \begin{tikzpicture}
      \node (image) at (0,0) {\includegraphics[width=1.0\linewidth]{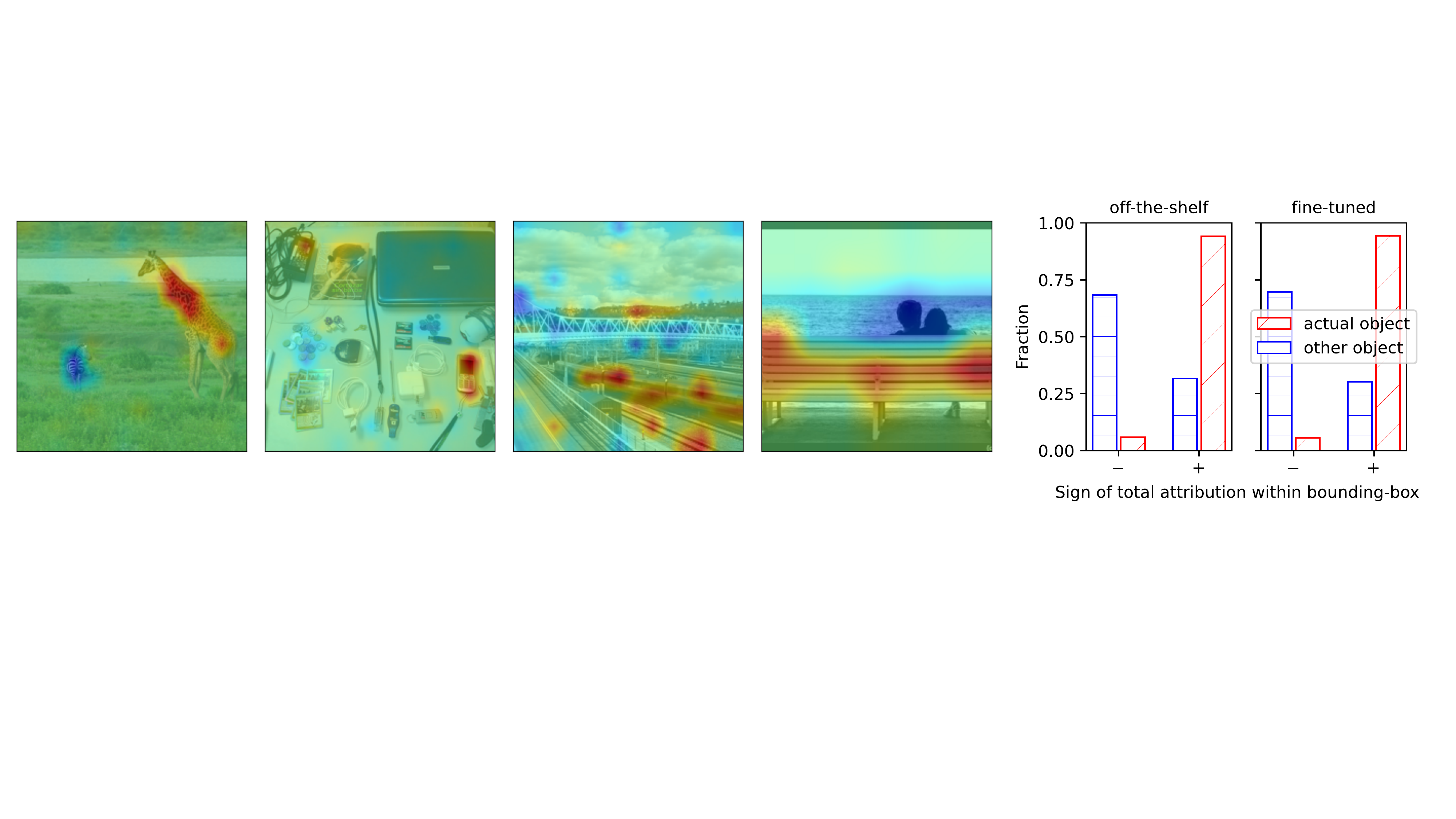}};
      \node[align=center] at (-6.8, -1.9) {\scriptsize A \adjustbox{bgcolor=yellow!100}{\strut giraffe} and a \\ \scriptsize \underline{zebra} checking \\ \scriptsize each other out.};
      \node[align=center] at (-4.0, -1.9) {\scriptsize A desk with \\ \scriptsize a \adjustbox{bgcolor=yellow!100}{\strut cellphone} \\ \scriptsize and \underline{pocket change}.};
      \node[align=center] at (-1.1, -1.9) {\scriptsize A long \underline{bridge} with \\ \scriptsize a \adjustbox{bgcolor=yellow!100}{\strut train} going \\ \scriptsize underneath.};
      \node[align=center] at (1.8, -1.9) {\scriptsize \underline{A couple} \\ \scriptsize \adjustbox{bgcolor=yellow!100}{\strut sitting on a bench} \\ \scriptsize looking at the sea.};
    \end{tikzpicture}
    \caption{(Left) \textbf{Examples of negative attributions for mismatches}. Attributions are for yellow selections in captions. Mismatching objects (underlined) receive negative attributions (blue). The histogram on the right shows the distribution over the sign of such cross-attributions.}
    \label{fig:neg_attr}
\end{figure}

\paragraph{Object Discrimination.} \label{sec:cross_attr}

We frequently observe that attributions between a given object in the text and a non-matching one in the image -- or vice versa -- are not only neutral but negative.
Figure \ref{fig:neg_attr} includes four explicit examples.
To systematically evaluate this effect, we sample instances from \gls{coco} that include at least two distinct object classes, each appearing exactly once in the image.
We then compute attributions between the two corresponding bounding-boxes and text spans and also across them, which we refer to as cross-attribution.
Attribution to the actual object's bounding-box is positive in \(94.1\%\) (\(94.1\%\)) of all cases, while cross-attributions to the other object are negative in \(68.4\%\) (\(69.7\%\)) of instances in the original (fine-tuned) model (cf. Figure \ref{fig:neg_attr} (right)).
This implies that \textbf{CLIP models do not only match corresponding objects across the input modes but can actively penalize mismatches by assigning them negative contributions}.

\begin{figure}[t]
    \centering
    \begin{tikzpicture}
      \node (image) at (0,0) {\includegraphics[width=0.9\textwidth]{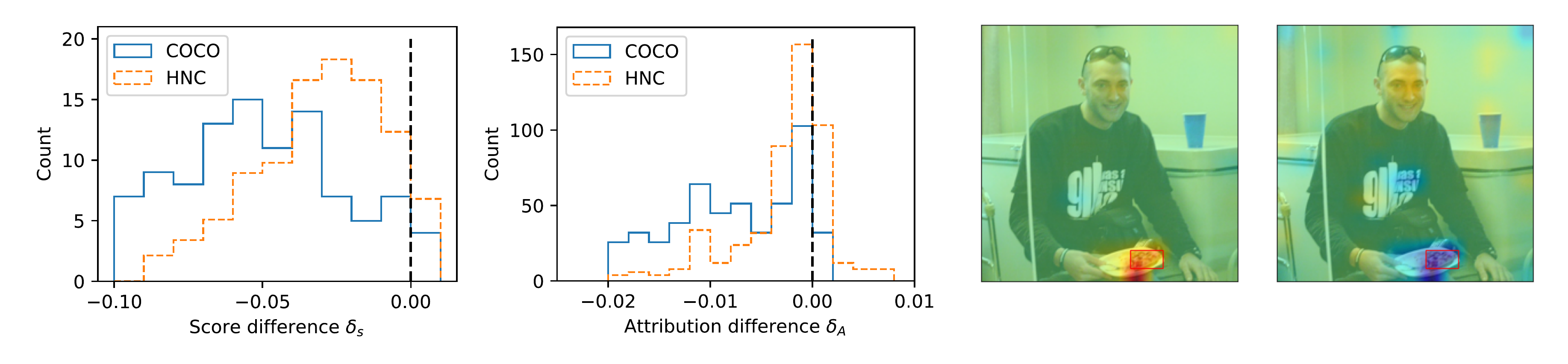}};
      \node[align=center] at (3.1, -1.4) {\scriptsize A man holding a \\ \scriptsize \adjustbox{bgcolor=yellow!100}{\strut slice of pizza}.};
      \node[align=center] at (5.9, -1.4) {\scriptsize A man holding a \\ \scriptsize \adjustbox{bgcolor=magenta!100}{\strut cup of coffee}.};
    \end{tikzpicture}
    \caption{\textbf{Attribution changes in hard negative captions}. (Left) Histograms for score (\(\delta_S\)) and attribution (\(\delta_A\)) differences. (Right) An example with the true caption on the left and a hard negative caption on the right. The true object is marked in yellow, and the replaced negative one in magenta.}
    \label{fig:hnc}
  \end{figure}

\paragraph{Hard negative captions.}
On the text side, it is straightforward to produce in-domain perturbations. 
We create hard negative captions that replace a single object in a positive caption with a reasonable but different object to receive a negative counterpart.
To this end, we leverage the automatic procedure by \citet{hnc} together with our simplified template (cf. Section \ref{sec:setting}) and additionally create a second resource from \gls{coco} by manually annotating a small yet high-quality evaluation sample of 100 image-caption pairs.\\
We check whether our negative captions actually result in a decrease of the predicted similarity score compared with their positive counterparts and define the difference as \(\delta_S\). It is negative in 95.2\% (89.1\%) of the \gls{coco} (\gls{hnc}) pairs.
We then compute attributions between the token range of the original or replaced object and the object bounding-box in the image and define the attribution difference as \(\delta_A\). It is also negative in 95.2\% (74.1\%) of the \gls{coco} (\gls{hnc}) examples.
Full histograms for \(\delta_S\) and \(\delta_A\) as well as an example for a change in attributions is included in Figure \ref{fig:hnc}.
These results show that \textbf{the model mostly reacts correctly to mistakes in captions and decreases the correspondence between affected image regions and caption spans}.

  \begin{figure*}[t]
    \centering
    \begin{tikzpicture}
      \node (image) at (0,0) {\includegraphics[width=\textwidth]{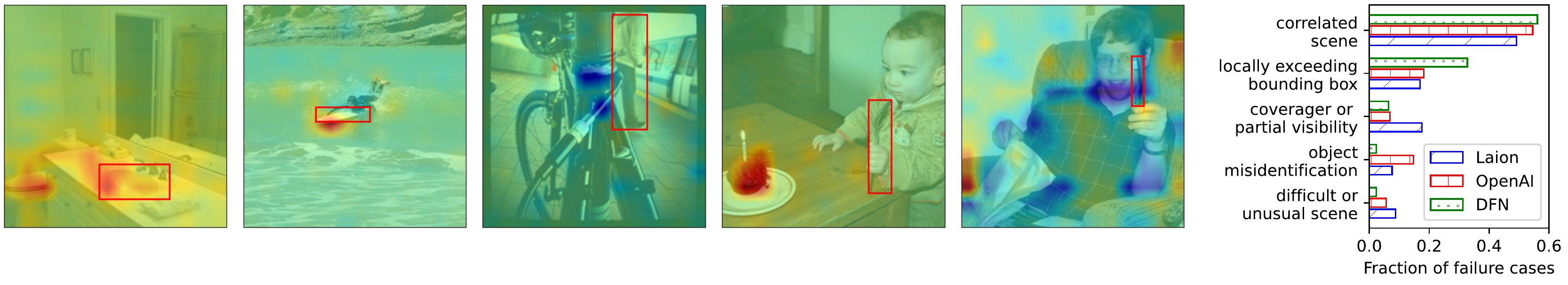}};
      \node[align=center] at (-7.1, -1.5) {\scriptsize A bathroom with \\ \scriptsize \adjustbox{bgcolor=yellow!100}{\strut sink} and toilet.};
      \node[align=center] at (-4.5, -1.5) {\scriptsize Someone on a \adjustbox{bgcolor=yellow!100}{\strut surf-} \\ \scriptsize \adjustbox{bgcolor=yellow!100}{\strut board} belly down.};
      \node[align=center] at (-2.0, -1.5) {\scriptsize A waiting \adjustbox{bgcolor=yellow!100}{\strut person} \\ \scriptsize behind a bicycle.};
      \node[align=center] at (0.5, -1.5) {\scriptsize A baby with a \\ \scriptsize \adjustbox{bgcolor=yellow!100}{\strut spoon} and cupcake.};
      \node[align=center] at (3.1, -1.5) {\scriptsize A guy holding \\ \scriptsize up a \adjustbox{bgcolor=yellow!100}{\strut toothbrush}.};
      \node[align=center] at (-7.1, 1.7) {\scriptsize correlation};
      \node[align=center] at (-4.5, 1.7) {\scriptsize locally exceeding};
      \node[align=center] at (-2.0, 1.7) {\scriptsize coverage};
      \node[align=center] at (0.5, 1.7) {\scriptsize misidentification};
      \node[align=center] at (3.0, 1.7) {\scriptsize unusual};
    \end{tikzpicture}
    \caption{\textbf{Examples for the five failure categories} that we can identify (left) and their relative occurrence in three models (right). More examples for all categories are in included in Figure \ref{fig:failures_examples}.}
    \label{fig:failure_main}
  \end{figure*}

\paragraph{Qualitative failure analysis.}
To identify cases where the models' grounding abilities are
systematically weak, we extract objects with \gls{pge} \(<0.2\) from the
\gls{coco} validation set and categorize them qualitatively.  For the
\laion, \openai, and \dfn models, this results in approximately $200$
image-caption pairs each.  \textbf{We can identify five major failure categories}: (1)
Visually correlated scenes like baseball courts, bathrooms,
offices, etc., (2) attributions locally exceeding bounding
  boxes, (3) coverage or partial visibility of objects, (4)
actual object misidentifications, and (5) difficult
  or unusual scenes.  Figure \ref{fig:failure_main} shows the
distribution among these categories and an example for each. More
examples are included in Figure \ref{fig:failures_examples}. Category
(1), correlated scenes, accounts for approximately half of
all failures in all three models, indicating that \gls{clip} models may struggle to differentiate between objects that commonly appear together.
\section{Discussion} \label{sec:discussion}

\paragraph{Interpretation of results and future work.}
While prior work has already established that \gls{clip} models can ground full text inputs onto images and vice versa \citep{gradeclip}, our second-order attributions take these insights a step further and show that this visual-linguistic correspondence is more fine-grained connecting individual parts of captions and images.\\
Our evidence for this intrinsic grounding ability to be significantly reduced on data outside the initial training domain complements recent efforts towards an understanding of \gls{clip}'s ostensible out-of-domain generalization \citep{xue}.
While \cite{fang} and \cite{nguyen} identified the training distribution as the critical component, \cite{mayilvahanan2024, clip_domain} recently showed that it must be assigned to domain contamination of web-scale training datasets and \gls{clip} models do \textit{not} actually generalize to unseen image domains, like renditions.\\
% Our results add to this that even in natural images the models can require explicit exposure to some object classes to establish a solid correspondence between the vision and language mode for the underlying concepts.\\
The finding that \gls{clip} models can actively assign negative contributions to mismatches reveals a non-trivial mechanism in their prediction computation.
The fact that this is not consistently the case and we also observe positive cross-attributions in correlated scenes like tennis courts, bathrooms, kitchens, streets, etc., however, is yet to be understood.
It suggests the contrastive objective may not provide sufficient supervision to learn to tell apart objects that frequently co-occur.
Future work should establish a detailed understanding of this phenomenon.
A solution may be to augment the training data with negatives targeting such correlations \citep{neg_clip, triplet_clip}.\\
Our baseline experiments show that analyzing interactions in \gls{clip} models is not trivial. Neither simplified gradient-based approaches (\gls{icam}), pair-wise embedding multiplication (\gls{itsm}) nor surrogate modeling (\gls{ilime}) are sufficient for the purpose.
% \gls{ilime} successfully fits text-side dependencies but cannot reliably capture interactions with image features.
It may still be possible to further enhance our method accounting for discrete text representations \citep{discrete_ig}, incorporating non-uniform interpolation \citep{non_uniform_ig} or integrating along non-linear paths \citep{guided_ig, ig2}.

\paragraph{Limitations.}
As stated explicitly in Equation \ref{eq:appr_attr}, our interaction attributions are an approximation.
Throughout this work, we attribute to intermediate representations of inputs, which is both efficient and informative \citep{emnlp}. 
In transformers, intermediate representations have undergone multiple contextualization steps and are technically not strictly tied to input features at a given position.
Finally, recently proven fundamental limitations of attribution methods urge caution in their interpretation especially regarding counterfactual conclusions about feature importance \citep{pnas}.\\
Despite these considerations, our consistent results on caption-image interactions across a variety of models and datasets provide strong empirical evidence for the evolvement of fine-grained inter-modal correspondence in \gls{clip} models through contrastive training.
While we must be more careful drawing definite conclusions about specific failure cases, we argue that explainability methods like ours can be used to formulate hypotheses about mistakes and biases in models.
We cannot regard them as guaranteed robust and faithful, but they provide insights that have the potential to improve models further.

\section{Conclusion}
We derive general second-order attributions in dual encoder architectures, enabling the attribution of similarity predictions onto interactions between input features.
Our method is applicable to any differentiable dual-encoder and requires no modifications of the initial model.
We believe it can also provide valuable insights into more complex relations between images and text \citep{visualgenome}, models for different modalities \citep{audioclip} and applications like retrieval \citep{xcolbert, vast}.
Our experiments with \gls{clip} models provide strong evidence for
them capturing fine-grained interactions between corresponding visual and
linguistic concepts despite their coarse contrastive objective.
At the same time, we also observe pronounced out-of-domain effects.
These results complement recent findings identifying limitations in the generalization capabilities of \gls{clip} \citep{clip_domain}.
Finally, an error analysis revealed that \gls{clip} models can struggle with covered or partially visible objects, unusual scenes, and correlated contexts like kitchens, offices, or sports courts.\\
By enabling the analysis of interactions between caption and image features, our approach contributes to an emerging interest in understanding higher-order dependencies in \gls{clip} models \citep{gandelsman25, ilime}, reaching beyond well-understood first-order effects \citep{gradeclip}.

\section*{Acknowledgements}\label{sec:acknowledgements}
Funded by Deutsche Forschungsgemeinschaft (DFG, German Research Foundation) under Germany’s Excellence Strategy - EXC 2075 – 390740016. We acknowledge the support by the Stuttgart Center for Simulation Science (SimTech).

% \newpage
\bibliography{references}
\bibliographystyle{tmlr}
\appendix
\section{Additional Examples} \label{apx:more_examples}

\begin{figure}[b!]
    \centering
    \begin{subfigure}[b]{0.23\textwidth}
    \centering
        \includegraphics[width=\textwidth]{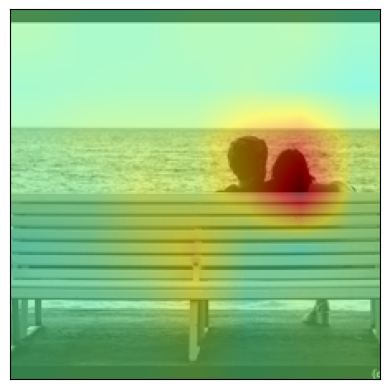}
        \subcaption{\adjustbox{bgcolor=yellow!100}{\strut A couple} sitting on a bench looking at the sea.}
    \end{subfigure}
    \begin{subfigure}[b]{0.23\textwidth}
        \centering
            \includegraphics[width=\textwidth]{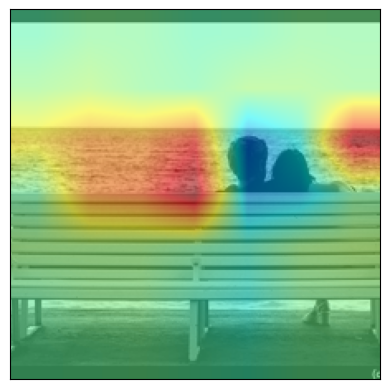}
            \subcaption{A couple sitting on a bench looking at \adjustbox{bgcolor=yellow!100}{\strut the sea}.}
    \end{subfigure}
    \hspace{0.45cm}
    \begin{subfigure}[b]{0.23\textwidth}
        \centering
        \includegraphics[width=\textwidth]{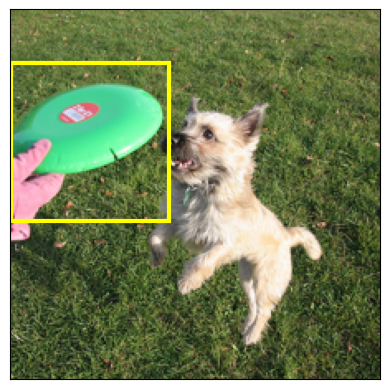}
        \subcaption{
            \adjustbox{bgcolor=blue!3}{\strut A} \adjustbox{bgcolor=blue!2}{\strut dog} \adjustbox{bgcolor=blue!2}{\strut is} \adjustbox{bgcolor=blue!18}{\strut jumping} \adjustbox{bgcolor=blue!15}{\strut for} \adjustbox{bgcolor=blue!1}{\strut a} \adjustbox{bgcolor=red!40}{\strut frisbee}.
        }
    \end{subfigure}
    \begin{subfigure}[b]{0.23\textwidth}
        \centering
        \includegraphics[width=\textwidth]{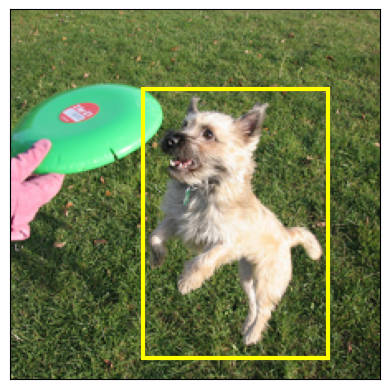}
        \subcaption{
            \adjustbox{bgcolor=red!5}{\strut A} \adjustbox{bgcolor=red!23}{\strut dog} \adjustbox{bgcolor=red!6}{\strut is} \adjustbox{bgcolor=blue!11}{\strut jumping} \adjustbox{bgcolor=blue!9}{\strut for} \adjustbox{bgcolor=blue!0}{\strut a} \adjustbox{bgcolor=blue!30}{\strut frisbee}.
        }
    \end{subfigure}
    \caption{Additional examples for inter-modal attributions of token-range selection with image projections (left) and bounding-box selection with caption projection (right). The visualization is identical to Figure \ref{fig:fist_vs_second_order}.}
    \label{fig:more_inter_modal}
\end{figure}

Figure \ref{fig:more_inter_modal} shows two additional examples for inter-modal attributions, one for text-span selection and image projection and one for bounding-box selection and caption projection.

\begin{figure*}
    \centering
    \vspace{-1.5cm}
    \begin{tikzpicture}
      \node (image) at (0,0) {\includegraphics[width=.7\textwidth]{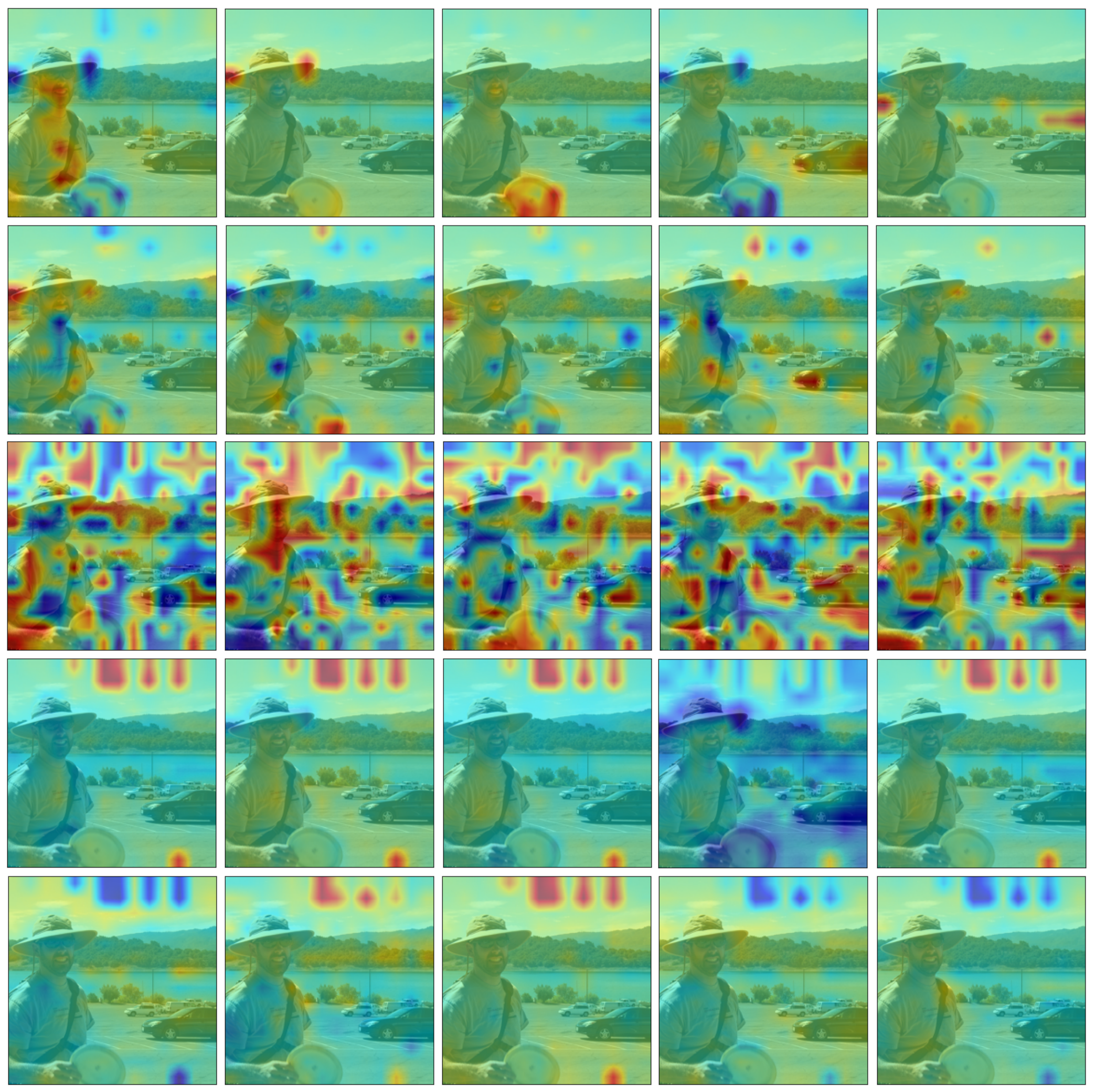}};
        \node at (-4.6,-5.9) {\scriptsize A  \adjustbox{bgcolor=yellow!100}{\strut guy}};
        \node at (-2.4,-5.9) {\scriptsize with a  \adjustbox{bgcolor=yellow!100}{\strut hat}};
        \node at (0,-5.9) {\scriptsize and a  \adjustbox{bgcolor=yellow!100}{\strut frisbee}};
        \node at (2.3,-5.9) {\scriptsize in front of a  \adjustbox{bgcolor=yellow!100}{\strut car}};
        \node at (4.6,-5.9) {\scriptsize by a  \adjustbox{bgcolor=yellow!100}{\strut lake}};
        \node[rotate=90] at (-6, 4.5) {\scriptsize Ours};
        \node[rotate=90] at (-6, 2.3) {\scriptsize ICAM};
        \node[rotate=90] at (-6, 0.0) {\scriptsize ILIME};
        \node[rotate=90] at (-6, -2.3) {\scriptsize \gls{itsm}$_{out}$};
        \node[rotate=90] at (-6, -4.5) {\scriptsize \gls{itsm}$_{hidden}$};
    \end{tikzpicture}
    \begin{tikzpicture}
      \node (image) at (0,0) {\scriptsize \includegraphics[width=.7\textwidth]{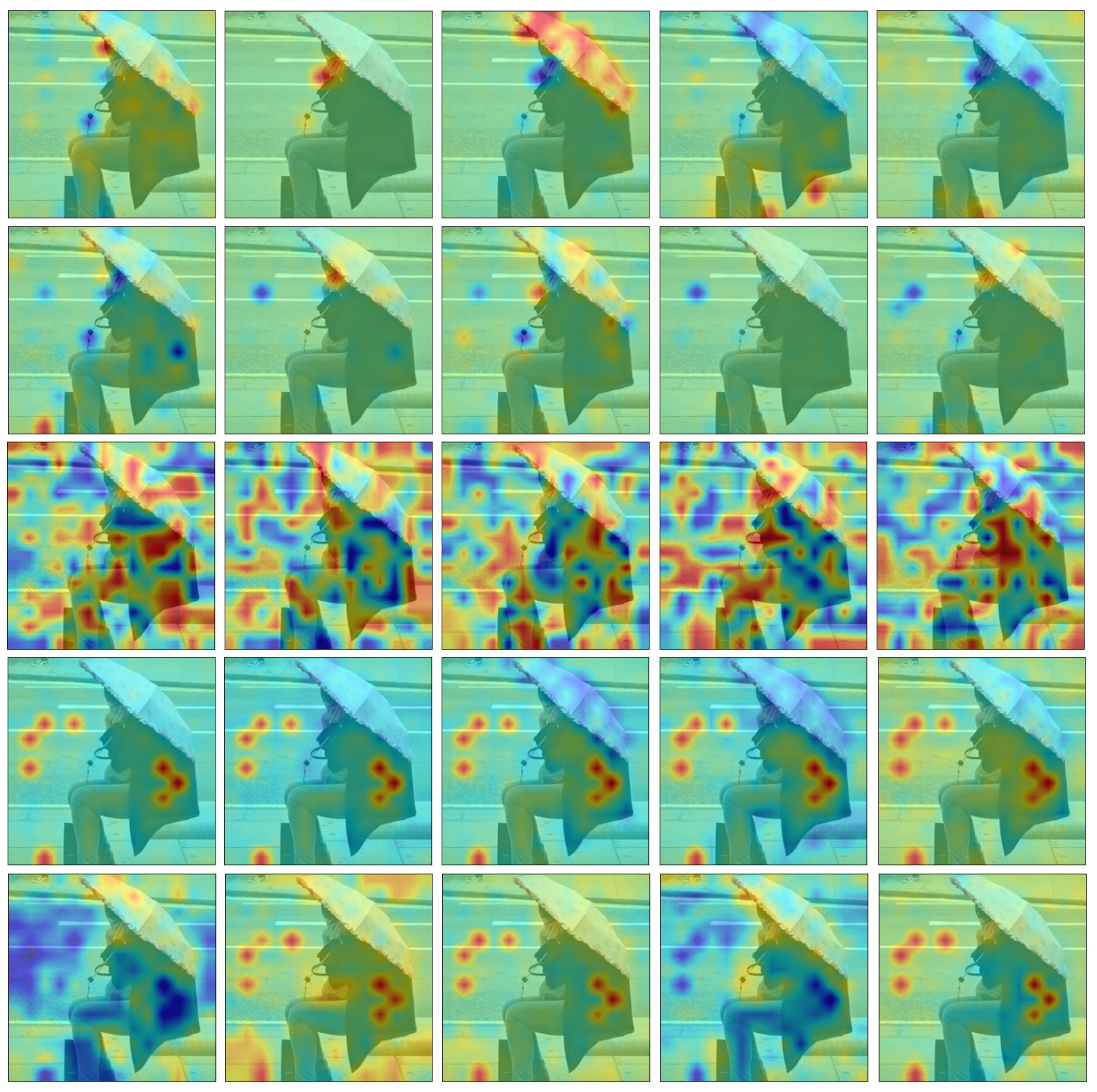}};
        \node at (-4.6,-5.9) {\scriptsize A \adjustbox{bgcolor=yellow!100}{\strut woman}};
        \node at (-2.4,-5.9) {\scriptsize \adjustbox{bgcolor=yellow!100}{\strut on the phone}};
        \node at (0,-5.9) {\scriptsize with an \adjustbox{bgcolor=yellow!100}{\strut umbrella}};
        \node at (2.3,-5.9) {\scriptsize sitting on a \adjustbox{bgcolor=yellow!100}{\strut bench}};
        \node at (4.6,-5.9) {\scriptsize in a \adjustbox{bgcolor=yellow!100}{\strut street}};
        \node[rotate=90] at (-6, 4.5) {\scriptsize Ours};
        \node[rotate=90] at (-6, 2.3) {\scriptsize ICAM};
        \node[rotate=90] at (-6, 0.0) {\scriptsize ILIME};
        \node[rotate=90] at (-6, -2.3) {\scriptsize \gls{itsm}$_{out}$};
        \node[rotate=90] at (-6, -4.5) {\scriptsize \gls{itsm}$_{hidden}$};
    \end{tikzpicture}
    \vspace{-.4cm}
    \caption{Qualitative comparison between our attributions, the \gls{icam}, \gls{ilime} and both \gls{itsm} variants. Heatmaps over images in a given column are for the marked parts of the captions in yellow below.}
    \label{fig:qualitative}
\end{figure*}

Figure \ref{fig:qualitative} shows a qualitative comparison between our attributions and the baselines described in Section \ref{sec:attr_eval}.

\begin{figure*}
    \centering
    \begin{tikzpicture}
      \node (image) at (0,0) {\includegraphics[width=\textwidth]{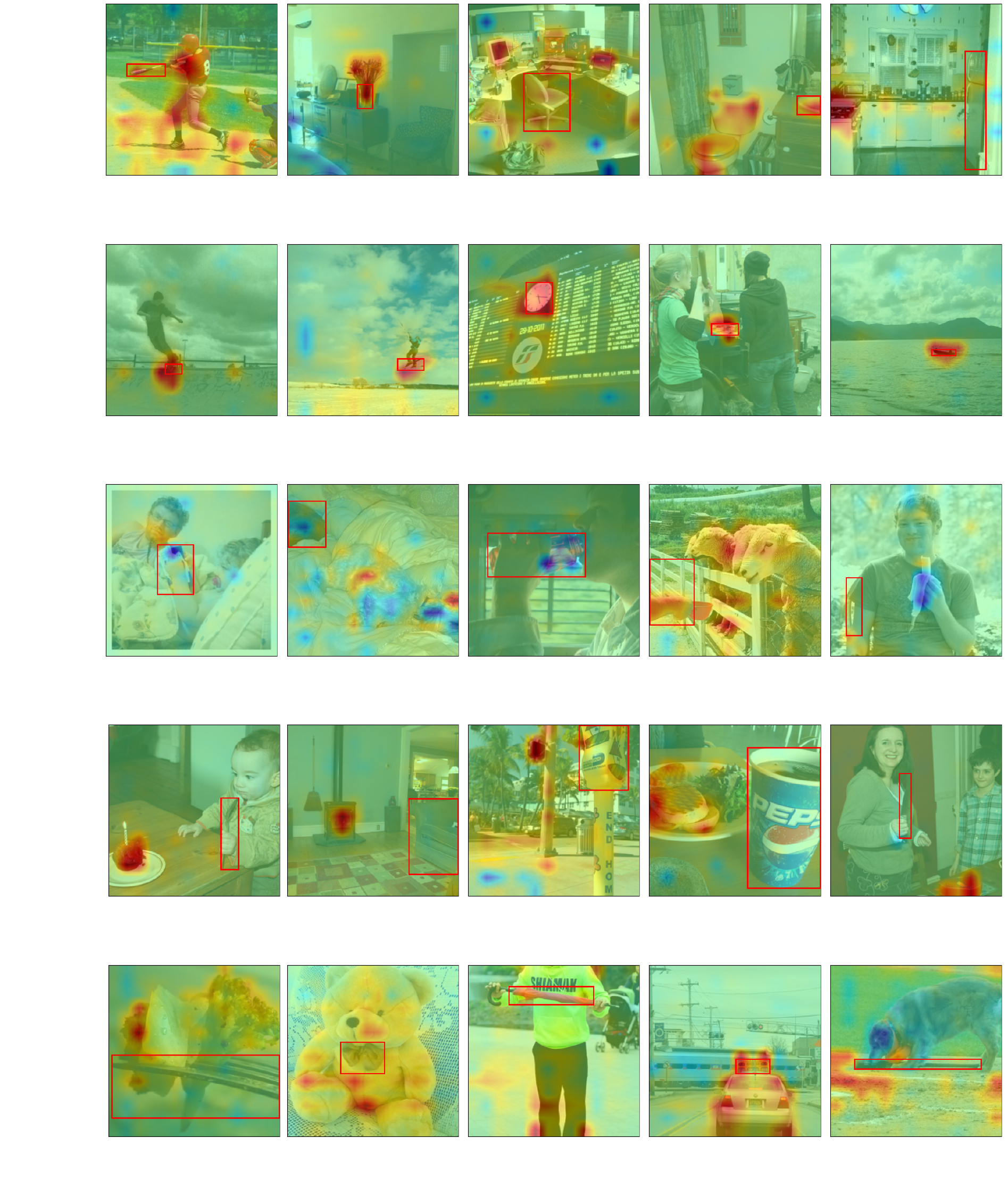}};
      % categories
      \node[rotate=90, align=center] at (-7.2, 8.2) {correlated \\ scene};
      \node[rotate=90, align=center] at (-7.2, 4.4) {locally exceeding \\ bounding boxes};
      \node[rotate=90, align=center] at (-7.2, .5) {coverage or \\ partial visibility};
      \node[rotate=90, align=center] at (-7.2, -3.5) {object \\ misidentification};
      \node[rotate=90, align=center] at (-7.2, -7.4) {difficult or \\ unusual scene};
      % captions for correlated scenes
      \node[align=center] at (-5.2, 6.45) {\scriptsize A baseball player \\ \scriptsize swinging a \adjustbox{bgcolor=yellow!100}{\strut bat}.};
      \node[align=center] at (-2.1, 6.45) {\scriptsize A living room with \\ \scriptsize a \adjustbox{bgcolor=yellow!100}{\strut vase} with flowers.};
      \node[align=center] at (0.8, 6.45) {\scriptsize An office with desk,\\ \scriptsize \adjustbox{bgcolor=yellow!100}{\strut chair} and laptop.};
      \node[align=center] at (3.7, 6.45) {\scriptsize Bathroom area, tub, \\ \scriptsize toilet and \adjustbox{bgcolor=yellow!100}{\strut sink}.};
      \node[align=center] at (6.75, 6.45) {\scriptsize Kitchen with stove, \\ \scriptsize fan and \adjustbox{bgcolor=yellow!100}{\strut refrigerator}.};
      % exceeding 
      \node[align=center] at (-5.2, 2.5) {\scriptsize A man riding a \\ \scriptsize \adjustbox{bgcolor=yellow!100}{\strut skateboard} up a ramp.};
      \node[align=center] at (-2.1, 2.5) {\scriptsize A person on \adjustbox{bgcolor=yellow!100}{\strut skies} \\ \scriptsize flies through the air.};
      \node[align=center] at (0.8, 2.5) {\scriptsize An airline status \\ \scriptsize board with a  \adjustbox{bgcolor=yellow!100}{\strut clock}.};
      \node[align=center] at (3.7, 2.5) {\scriptsize A person holds a \\ \scriptsize \adjustbox{bgcolor=yellow!100}{\strut pizza} on a peel.};
      \node[align=center] at (6.75, 2.5) {\scriptsize Two people sitting on \\ \scriptsize a small \adjustbox{bgcolor=yellow!100}{\strut boat} floating.};
      % coverage
      \node[align=center] at (-5.2, -1.4) {\scriptsize A child in bed looking \\ \scriptsize at a picutre \adjustbox{bgcolor=yellow!100}{\strut book}.};
      \node[align=center] at (-2.1, -1.4) {\scriptsize A dog next to a \adjustbox{bgcolor=yellow!100}{\strut cat} \\ \scriptsize lying on a bed.};
      \node[align=center] at (0.8, -1.4) {\scriptsize A man drinking \\ \scriptsize from a red  \adjustbox{bgcolor=yellow!100}{\strut bottle}.};
      \node[align=center] at (3.7, -1.4) {\scriptsize A \adjustbox{bgcolor=yellow!100}{\strut person} feeding \\ \scriptsize sheep behind a fence.};
      \node[align=center] at (6.75, -1.4) {\scriptsize A man sitting in a \adjustbox{bgcolor=yellow!100}{\strut chair} \\ \scriptsize eating a banana.};
      % misidentification
      \node[align=center] at (-5.2, -5.35) {\scriptsize A baby with a \adjustbox{bgcolor=yellow!100}{\strut spoon} \\ \scriptsize looking at a cupcake.};
      \node[align=center] at (-2.1, -5.35) {\scriptsize A living room with a \\ \scriptsize \adjustbox{bgcolor=yellow!100}{\strut couch} and a rug.};
      \node[align=center] at (0.8, -5.35) {\scriptsize A \adjustbox{bgcolor=yellow!100}{\strut parking meter} \\ \scriptsize decorated with a house.};
      \node[align=center] at (3.7, -5.35) {\scriptsize A plate of food next to\\ \scriptsize a \adjustbox{bgcolor=yellow!100}{\strut cup} of pepsi.};
      \node[align=center] at (6.75, -5.35) {\scriptsize A woman with a \adjustbox{bgcolor=yellow!100}{\strut knife} \\ \scriptsize about to cut some cake.};
      % difficult
      \node[align=center] at (-5.2, -9.3) {\scriptsize A closeup of a \adjustbox{bgcolor=yellow!100}{\strut fork} \\ \scriptsize holding some broccoli.};
      \node[align=center] at (-2.1, -9.3) {\scriptsize A tan teddy bear \\ \scriptsize wearing a bow \adjustbox{bgcolor=yellow!100}{\strut tie}.};
      \node[align=center] at (0.8, -9.3) {\scriptsize Someone holding an\\ \scriptsize \adjustbox{bgcolor=yellow!100}{\strut umbrella} on a sidewalk.};
      \node[align=center] at (3.7, -9.3) {\scriptsize Cars and a \adjustbox{bgcolor=yellow!100}{\strut bus} \\ \scriptsize stopped for a train.};
      \node[align=center] at (6.75, -9.3) {\scriptsize A dog that is\\ \scriptsize sniffing a baseball  \adjustbox{bgcolor=yellow!100}{\strut bat }.};
    \end{tikzpicture}
    \caption{Five examples for each of the five identified failure categories as described in Section \ref{sec:model_analysis}.}
    \label{fig:failures_examples}
\end{figure*}

Figure \ref{fig:failures_examples} shows five examples for each of the five failure categories that we identified in Section \ref{sec:model_analysis} under \textit{Qualitative failure analysis}.

\begin{figure}[tb!]
    \centering
    \begin{subfigure}[b]{0.46\textwidth}
        \includegraphics[width=\textwidth]{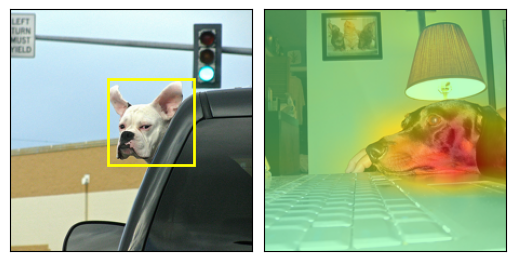}
    \end{subfigure}
    \hspace{0.45cm}
    \begin{subfigure}[b]{0.46\textwidth}
        \includegraphics[width=\textwidth]{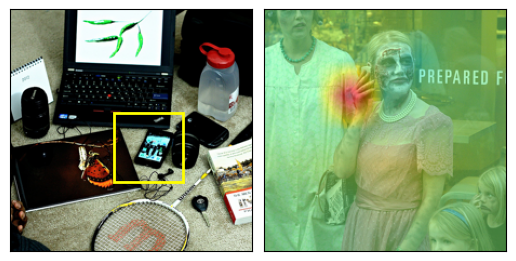}
    \end{subfigure}
    \caption{Image-image attributions between the yellow bounding-box in the left image and the one to its right as described in Section \ref{sec:method}. Visualisation is identical to Figure \ref{fig:intra-modal} (right)}
    \label{fig:more_img_img}
\end{figure}

\section{Intra-modal attributions} \label{apx:intra-modal}.
This section describes intra-modal model attributions for text- or image-pairs exemplified in Figure \ref{fig:intra-modal}
For text-text attributions, after summation over embedding
dimensions, the attributions take the form of an \(S_1 \!\times\! S_2\) dimensional matrix, with \(S_1\) and \(S_2\) being token sequence lengths of the two texts. 
For image-image pairs, attribution tensors become four dimensional
taking the shape \((H \!\times\! W)_1 \!\times\! (H \!\times\! W)_2\), containing a contribution for every pair of two patches from either image. 
Figure \ref{fig:more_img_img} includes additional examples.

\section{Extended results} \label{apx:add_results}

\begin{figure}[tb!]
    \centering
    \begin{subfigure}[b]{0.42\textwidth}
        \includegraphics[width=\textwidth]{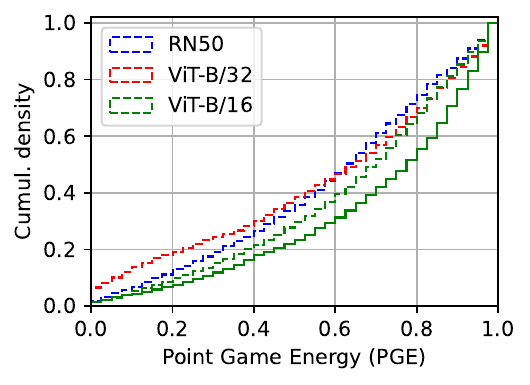}
    \end{subfigure}
    \begin{subfigure}[b]{0.42\textwidth}
        \includegraphics[width=\textwidth]{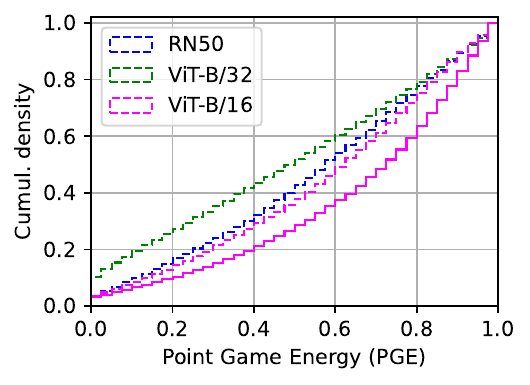}
    \end{subfigure}
    \caption{Cumulative \gls{pge}-distribution plots of the unmodified (dashed) / fined-tuned (solid) \openai~models on \gls{coco} (left) and \gls{flickr} (right) dataset as described in Section \ref{sec:bbox_attr}. }
    \label{fig:add_cumul_bbox_attr}
\end{figure}

\begin{figure}[tb!]
    \centering
    \begin{subfigure}[b]{0.42\textwidth}
        \includegraphics[width=\textwidth]{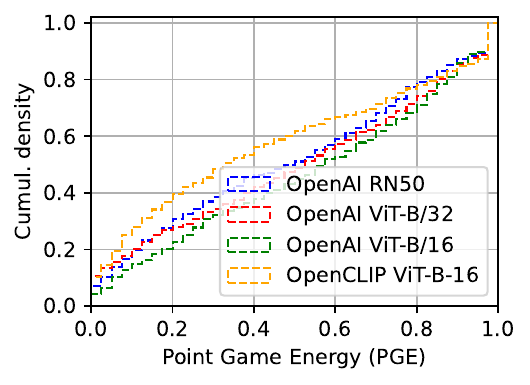}
    \end{subfigure}
    \begin{subfigure}[b]{0.42\textwidth}
        \includegraphics[width=\textwidth]{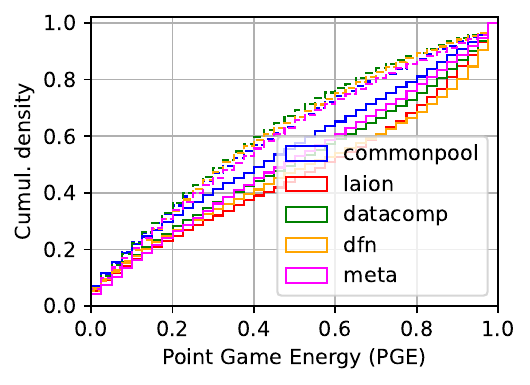}
    \end{subfigure}
    \caption{Cumulative \gls{pge}-distribution plots for the different models on HNC (left) and the unmodified (dashed) / fine-tuned (solid) \openclip~models on \gls{flickr} (right) as described in Section \ref{sec:bbox_attr}.}
    \label{fig:add_cumul_bbox_attr_2}
\end{figure}

Table \ref{tab:openai_bbox_attr} shows full results for our Point-Game evaluation on different \openai~models. 
Next to the ViT-B/16 architecture, we also evaluate the RN50 and ViT-B/32 variants.
Table \ref{tab:openclip_bbox_attr} includes the full evaluation for all \openclip~models. 
In addtion to the median \gls{pge} (mPGE), in these tables we also report cumulative \gls{pge} densitites for the \(80^{th}\) percentile (\gls{pge}>0.8).
Full cumulative \gls{pge}-histograms for additional models are included in Figures \ref{fig:add_cumul_bbox_attr} and \ref{fig:add_cumul_bbox_attr_2}.

\begin{figure}
    \centering
    \begin{subfigure}[T]{0.42\textwidth}
        \includegraphics[width=\textwidth]{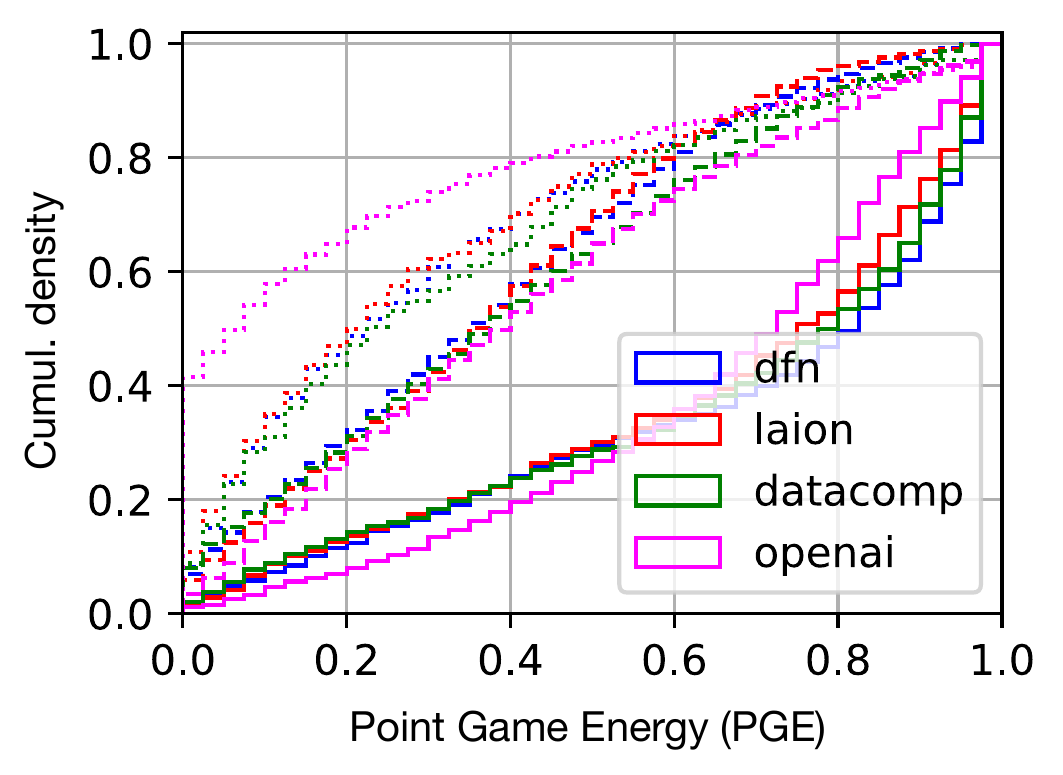}
    \end{subfigure}
    \caption{Cumulative PGE-distributions for our baseline experiment in Section \ref{sec:attr_eval}. Our method is in solid, InteractionCAM is dashed and \gls{itsm}$_{out}$ is dotted. \gls{itsm}$_{hidden}$ is excluded for an uncluttered visualization.}
    \label{fig:pge_hist_ablation}
\end{figure}

\begin{figure}[htb!]
    \centering
    \begin{subfigure}[b]{0.42\textwidth}
        \includegraphics[width=\textwidth]{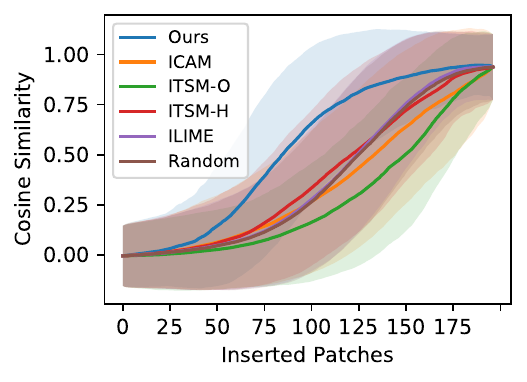}
        \subcaption{Conditional Image Insertion.}
    \end{subfigure}
    \begin{subfigure}[b]{0.42\textwidth}
        \includegraphics[width=\textwidth]{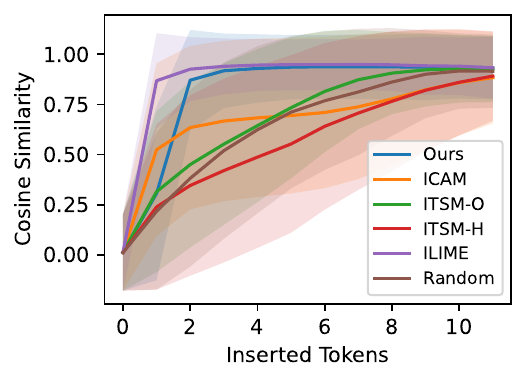}
        \subcaption{Conditional Text Insertion.}
    \end{subfigure}
    \begin{subfigure}[b]{0.42\textwidth}
        \includegraphics[width=\textwidth]{plots/perturbs_deletion_cumul_topk_ViT-B-16_laion2b_s34b_b88k_coco_image.pdf}
        \subcaption{Conditional Image Deletion.}
    \end{subfigure}
    \begin{subfigure}[b]{0.42\textwidth}
        \centering
        \includegraphics[width=\textwidth]{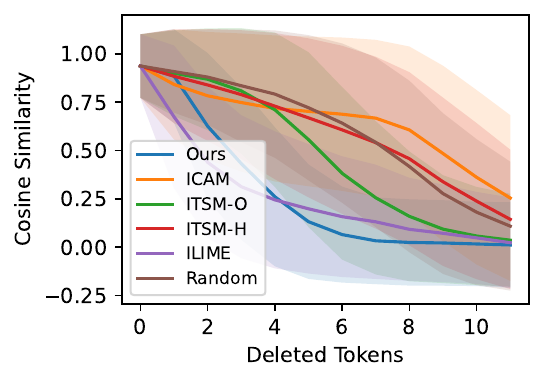}
        \subcaption{Conditional Text Deletion.}
    \end{subfigure}
    \caption{Average change in similarity score upon conditional insertion and deletion performed on either the caption or the image using a ViT-B-16 model pretrained on \laion. Confidence intervals are standard deviations over the evaluation dataset. Table \ref{tab:perturbs} summarizes the AUC of these plots.}
    \label{app:fig:conditional_insert_deletion_laion}
\end{figure}

\begin{figure}[htb!]
    \centering
    \begin{subfigure}[b]{0.42\textwidth}
        \includegraphics[width=\textwidth]{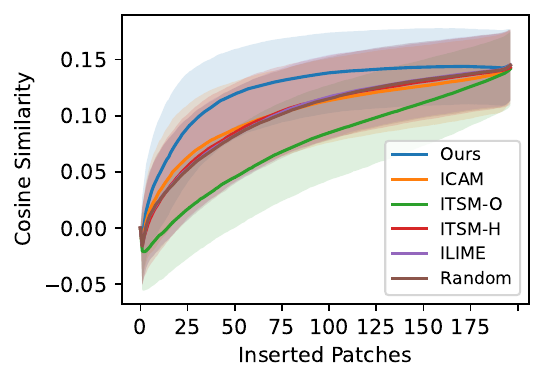}
        \subcaption{Conditional Image Insertion.}
    \end{subfigure}
    \begin{subfigure}[b]{0.42\textwidth}
        \includegraphics[width=\textwidth]{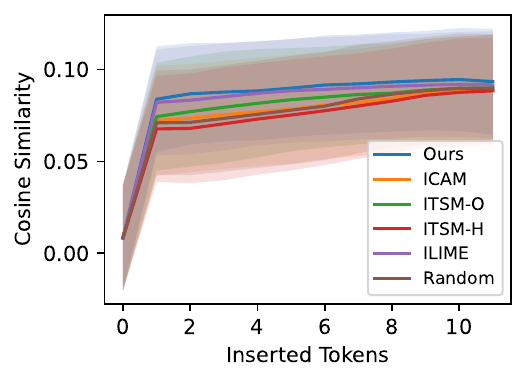}
        \subcaption{Conditional Text Insertion.}
    \end{subfigure}
    \begin{subfigure}[b]{0.42\textwidth}
        \includegraphics[width=\textwidth]{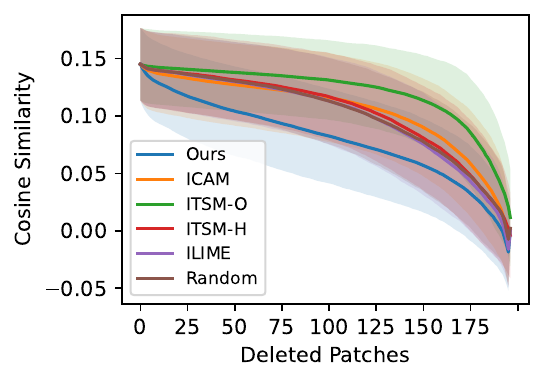}
        \subcaption{Conditional Image Deletion.}
    \end{subfigure}
    \begin{subfigure}[b]{0.42\textwidth}
        \centering
        \includegraphics[width=\textwidth]{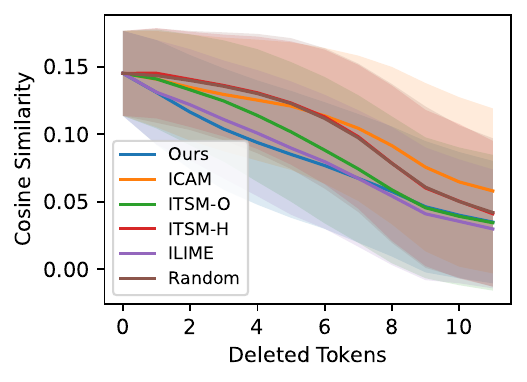}
        \subcaption{Conditional Text Deletion.}
    \end{subfigure}
    \caption{Conditional insertion and deletion performed on either the caption or the image using the original ViT-B/16 model by \openai~without fine-tuning.}
    \label{app:fig:conditional_insert_deletion_openai}
\end{figure}

Table \ref{tab:full_baselines} presents full results of our Point-Game baseline experiments extending Section \ref{sec:attr_eval}.
Corresponding cumulative densities of the PGE-metric are shown in Figure \ref{fig:pge_hist_ablation}.
Figures \ref{app:fig:conditional_insert_deletion_laion} and \ref{app:fig:conditional_insert_deletion_openai} show the plots of the conditional insertion and deletion experiments for the \openclip~\laion~model and the original \openai~model, respectively. The corresponding \gls{auc} values are contained in Table \ref{tab:perturbs}.

\begin{figure}[tb!]
    \centering
    \begin{subfigure}[b]{0.9\textwidth}
        \includegraphics[width=\textwidth]{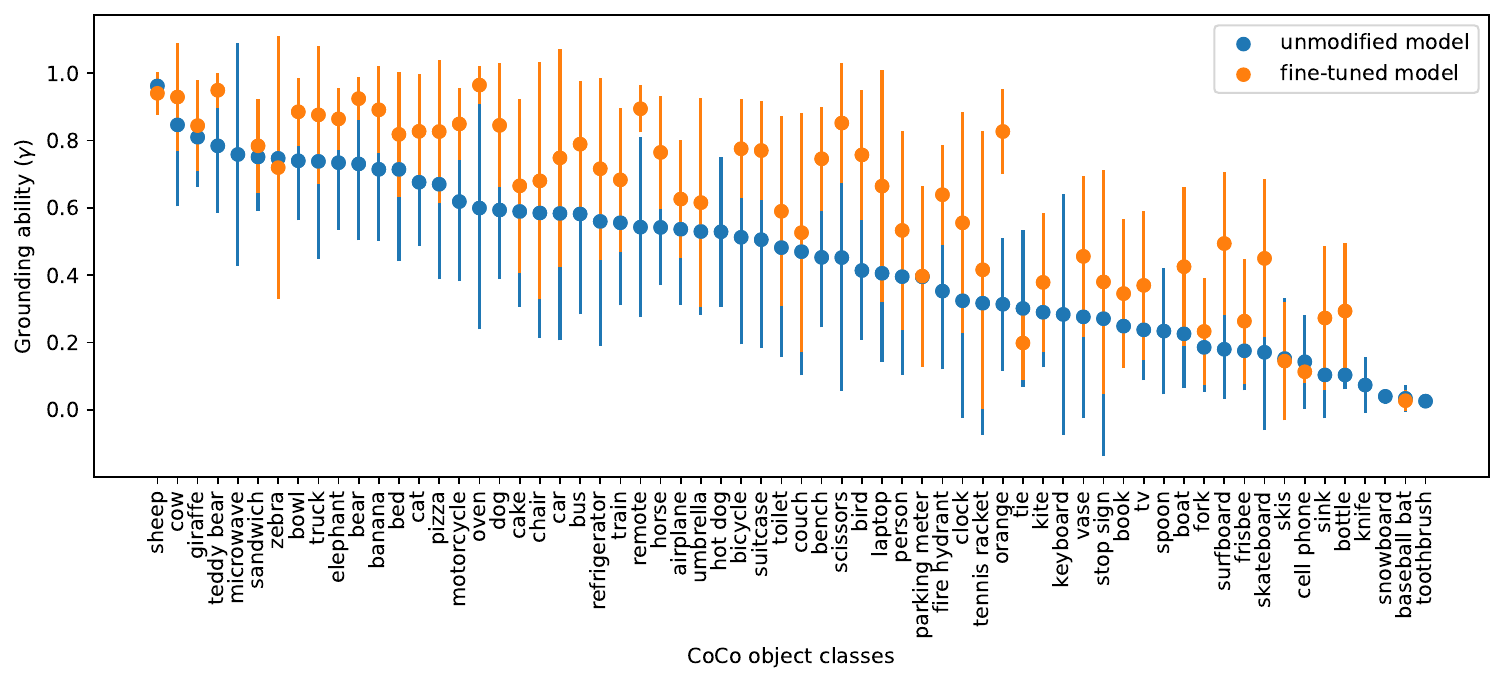}
    \end{subfigure}
    \hfill
    \begin{subfigure}[b]{0.9\textwidth}
        \includegraphics[width=\textwidth]{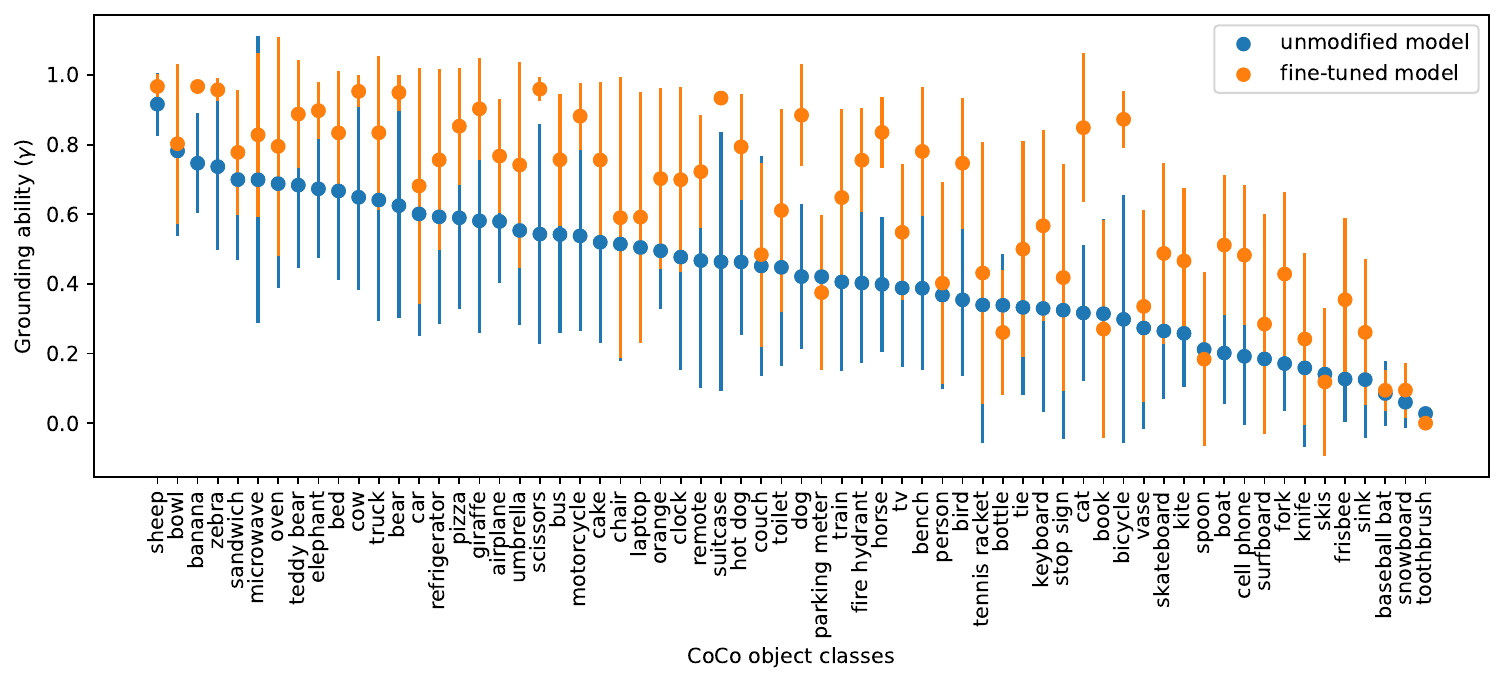}
    \end{subfigure}
    \caption{Class-wise \gls{pge}-evaluation for the \openclip~\laion~(top) and \datacomp~(bottom) models before and after in-domain fine-tuning as discussed in Section \ref{sec:bbox_attr}.}
    \label{fig:add_class_gamma}
\end{figure}

Figure \ref{fig:add_class_gamma} extends the class-wise PGE-evaluation from Section \ref{sec:model_analysis} to the \openclip~\dfn~and \datacomp~models.

\begin{table}[tb!]
    \centering
    \caption{Summary of the Point-Game evaluation for different \gls{clip}~models by \openai~as described in Section~\ref{sec:bbox_attr}. \textit{Model} refers to the investigated architecture, \textit{Tuning} is whether the model was fine-tuned on the train split of the respective dataset. Best overall results are in bold, best results of unmodified models are underlined.}
    \label{tab:openai_bbox_attr}
    \vspace{.2cm}
    \resizebox{1.0\textwidth}{!}{
    \begin{tabular}{rcccccccccc}
        \toprule
        \multicolumn{2}{c}{} & \multicolumn{3}{c}{\gls{coco}} & \multicolumn{3}{c}{HNC} & \multicolumn{3}{c}{\gls{flickr}} \\
        \cmidrule(r){3-5} \cmidrule(r){6-8} \cmidrule(r){9-11}
        \textbf{Model} & \textbf{Tuning} & mPGE & PGE>0.8 & PGA & mPGE & PGE>0.8 & PGA & mPGE & PGE>0.8 & PGA \\
        \midrule
        RN50 & No & 66.3 & 28.8 & 76.9 & 50.1 & 22.6 & 61.8 & 60.1 & 25.5 & 71.2 \\
        %RN101 & 54.0 & 24.8 & 72.5 & 44.2 & 22.8 & 60.6& & & \\
        ViT-B/32 & No & 63.5 & 33.3 & 69.1 & 52.8 & 28.5 & 58.5 & 50.4 & 23.4 & 58.1 \\ % final_flickr_results
        ViT-B/16 & No & \underline{72.3} & \underline{35.7} & \underline{79.0} & \underline{57.0} & \underline{31.7} & \underline{65.0} & \underline{64.4} & \underline{28.4} & \underline{72.1} \\
        ViT-B-16 & Yes & \textbf{78.0} & \textbf{48.4} & \textbf{82.9} & - & - & - & \textbf{73.4} & \textbf{40.7} & \textbf{79.0} \\
        \bottomrule
    \end{tabular}
    }
\end{table}

\begin{table}[tb!]
    \centering
    \caption{Summary of the Point-Game evaluation for all \openclip~models on \gls{coco} and Flickr30k. The \textit{Training} column refers to the dataset the model was initially trained on, \textit{Tuning} is whether the model was additionally fine-tuned on the train-split of the respective evaluation dataset. All models implement the ViT-B-16 architecture except Meta-\gls{clip} that uses quickgelu activations. Best overall results are in bold, best results for unmodified models are underlined.}
    \label{tab:openclip_bbox_attr}
    \vspace{.2cm}
    \resizebox{.8\textwidth}{!}{
    \begin{tabular}{rccccccc}
        \toprule
        \multicolumn{2}{c}{} & \multicolumn{3}{c}{COCO} & \multicolumn{3}{c}{\gls{flickr}} \\
        \cmidrule(r){3-5} \cmidrule(r){6-8} 
        \textbf{Training} & \textbf{Tuning} & mPGE & PGE>0.8 & PGA & mPGE & PGE>0.8 & PGA \\
        \midrule
        \multirow{2}{*}{\laion} & No & \underline{49.4} & 22.0 & \underline{63.3} & \underline{38.2} & \underline{15.9} & 52.0 \\
         & Yes & 71.1 & 47.3 & \textbf{83.2} & \textbf{54.6} & 30.6 & \textbf{61.8} \\
        \multirow{2}{*}{\commonpool} & No & 43.0 & 18.2 & 58.8 & 36.7 & 15.5 & \underline{53.0} \\
         & Yes & 57.7 & 28.7 & 67.1 & 44.6 & 20.8 & 56.2 \\
         \multirow{2}{*}{\datacomp} & No & 38.5 & 14.6 & 56.0 & 32.8 & 11.8 & 48.9 \\
         & Yes & \textbf{72.4} & 50.0 & 75.1 & 50.7 & 27.3 & 56.0 \\
         \multirow{2}{*}{DFN} & No & 46.5 & \underline{19.6} & 54.3 & 35.4 & 12.3 & 43.3 \\
         & Yes & 71.4 & \textbf{53.3} & 74.6 & 53.1 & \textbf{33.5} & 58.3 \\
         \multirow{2}{*}{Meta-\gls{clip}} & No & 44.2 & 16.8 & 52.3 & 37.0 & 14.5 & 46.4 \\
         & Yes & 57.5 & 49.8 & 77.1 & 49.2 & 24.1 & 57.2 \\
        \bottomrule
    \end{tabular}
    }
\end{table}

\begin{table}[tb!]
    \centering
        \begin{tabular}{rlcccc}
            \toprule
             \multicolumn{2}{c}{} & \multicolumn{2}{c}{\gls{coco}} & \multicolumn{2}{c}{Flickr30k} \\
            \cmidrule(r){3-4} \cmidrule(r){5-6}
            \textbf{Training} & \textbf{Method} & mPGE & PGA & mPGE & PGA \\
            \midrule
            \multirow{4}{*}{OpenAI} & \gls{itsm}$_\text{out}$ & 18.1 & 21.4 & 19.5 & 23.3 \\
            & \gls{itsm}$_\text{hidden}$ & 29.8 & 38.1 & 29.5 & 37.4 \\
            & ILIME & 27.9 & 34.9 & 25.8 & 33.1 \\
            & ICAM & 38.6 & 54.6 & 33.5 & 51.4 \\
            & Ours & \textbf{72.3} & \textbf{79.0} & \textbf{64.4} & \textbf{72.1} \\
            % Werte stimmen noch nicht
            \hline
            \multirow{4}{*}{\laion~(tuned)} & \gls{itsm}$_\text{out}$ & 22.8 & 30.3 & 24.5 & 28.7 \\
            & \gls{itsm}$_\text{hidden}$ & 30.5 & 34.6 & 28.8 & 36.6 \\
            & ILIME & 28.8 & 37.8 & 25.8 & 34.5 \\
            & ICAM & 32.5 & 58.4 & 33.5 & 51.4 \\
            & Ours & \textbf{71.2} & \textbf{83.2} & \textbf{56.3} & \textbf{63.6} \\
            \hline
            \multirow{4}{*}{\dfn~(tuned)} & \gls{itsm}$_\text{out}$ & 24.2 & 34.5 & 25.1 & 31.4 \\
            & \gls{itsm}$_\text{hidden}$ & 27.4 & 23.0 & 27.7 & 36.5 \\
            & ILIME & 27.9 & 39.2 & 25.7 & 33.5 \\
            & ICAM & 33.3 & 46.5 & 24.2 & 42.2 \\
            & Ours & \textbf{71.4} & \textbf{74.6} & \textbf{53.1} & \textbf{58.3} \\
            \hline
            \multirow{4}{*}{\datacomp~(tuned)} & \gls{itsm}$_\text{out}$ & 25.5 & 38.7 & 26.5 & 33.9 \\
            & \gls{itsm}$_\text{hidden}$ & 35.0 & 42.3 & 22.6 & 28.4 \\
            & ILIME & 28.4 & 39.3 & 25.7 & 34.1 \\
            & ICAM & 36.9 & 49.5 & 23.2 & 37.3 \\
            & Ours & \textbf{72.4} & \textbf{75.1} & \textbf{50.7} & \textbf{60.0} \\
            \bottomrule
        \end{tabular}
    \caption{PGE-evaluation results of our method compared against the \gls{itsm} and InteractionCAM (ICAM) baselines for different models as described in Section \ref{sec:attr_eval} under \textit{Object localization}. Best results for every model are in bold.}
    \label{tab:full_baselines}
\end{table}

\section{Additional Experiments} \label{apx:add_exp}

\paragraph{Approximation Error.}

\begin{figure}[tb!]
    \centering
    \includegraphics[width=0.42\linewidth]{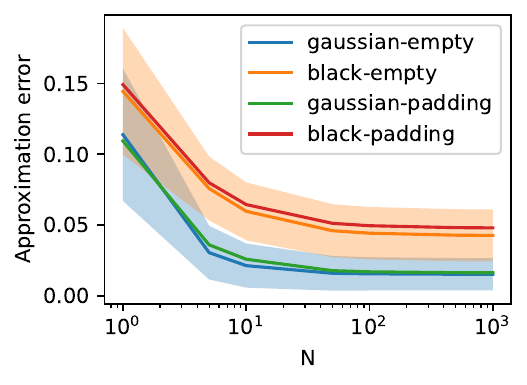}
    \caption{Approximation errors for different reference choices as a function of the number of integration steps \(N\). The image references are abbreviated as 'gaussian' for gaussian noise and 'black' for the black image. Text references are 'padding' and 'empty' for a padding sequence and the empty sequence, respectively. Exemplatory standard deviations over the evaluation sample are shown as shades of the respective plots.}
    \label{fig:appr_err}
\end{figure}

In Section \ref{sec:method}, we have shown the equality between Eq. \ref{eq:ansatz} and Eq. \ref{eq:attr_mat}.
The only approximation affecting this equality is the numerical integration in Eq. \ref{eq:int_jac} to calculate the two \textit{integrated Jacobians} \(\mathbf{J}^g\) and \(\mathbf{J}^h\) by a sum over \(N\) bins.
We can evaluate how good this approximation is by explicitly calculating the four similarity predictions between the references and inputs \(f(\mathbf{a}, \mathbf{b})\), \(f(\mathbf{r}_a, \mathbf{b})\), \(f(\mathbf{a}, \mathbf{r}_b)\) and \(f(\mathbf{r}_a, \mathbf{r}_b)\), as well as the attribution matrix \(\mathbf{A}\).
The \textit{approximation error} can then be defined as the absolute difference between Eq. \ref{eq:ansatz} and Eq. \ref{eq:attr_mat}.
In Figure \ref{fig:appr_err}, we plot this error as a function of different magnitudes for \(N\).
For larger \(N\), it converges as expected.

\begin{figure*}[tb!]
    \centering
    \begin{tikzpicture}
      \node (image) at (0,0) {\includegraphics[width=\textwidth]{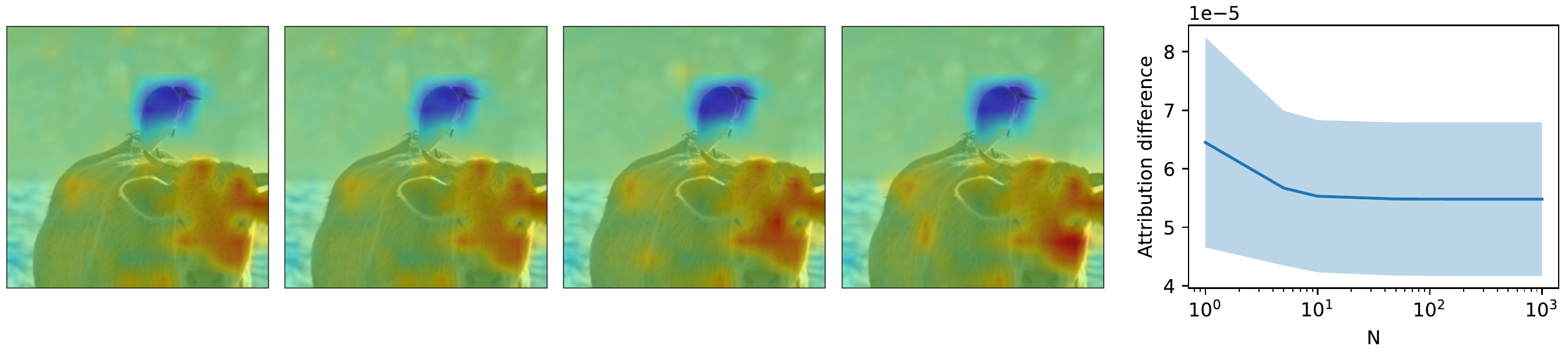}};
      \node[align=center] at (-2.5, -1.5) {\scriptsize Bird sitting atop \adjustbox{bgcolor=yellow!100}{\strut cattle} on snow covered field.};
      \node[align=center] at (-6.8, 1.8) {\scriptsize black-padding};
      \node[align=center] at (-3.9, 1.8) {\scriptsize black-empty};
      \node[align=center] at (-1.0, 1.8) {\scriptsize gaussian-padding};
      \node[align=center] at (2.0, 1.8) {\scriptsize gaussian-empty};
    \end{tikzpicture}
    \caption{(Left) Attribution differences for all four combinations of references as indicated above the images for the yellow selection in the caption below. (Right) The mean absolute attribution difference between a \textit{black-padding} and a \textit{gaussian-empty} reference combination as a function of the number of approximation steps \(N\) and its standard deviation over the evaluation smaple (blue shade).}
    \label{fig:ref_diff}
\end{figure*}

\paragraph{Choice of references.}
The choice of the references \(\mathbf{r}_a\) and \(\mathbf{r}_b\) is ambiguous as long as they are uninformative. 
We try different options and evaluate their approximation errors as defined above.
For the image input, we use a black image and zero-centered gaussian noise.
For the text input, we use a sequence of padding tokens and the empty sequence consisting only of the \texttt{CLS} and \texttt{EOS} tokens.
Figure \ref{fig:appr_err} includes approximation errors for all four combinations of these references as a function of \(N\).
Combinations with gaussian noise for the image reference appear to converge slightly faster than the black image.\\
For different references, attributions can vary slightly. 
However, these differences are small, even for large objects like the cow in Figure \ref{fig:ref_diff} (left), for which attributions tend to spread out.
The absolute difference of attributions between any two combinations of references is on the order of \(10^{-5}\) and tends to decrease for larger \(N\).
The plot in Figure \ref{fig:ref_diff} (right) shows this for the difference between a \textit{black-padding} and \textit{gaussian-empty} reference combination.

% \section{Alternative Figure 1}

% \begin{figure*}[tb!]
%     \includegraphics[width=\textwidth]{plots/first_vs_second_cubes.pdf}
%     \caption{Alternative illustration of the three dimensional character of our interaction attributions (left column) vs. first-order feature attributions (right column). Our attributions are a 3d volume with the image dimensions (H, W) and the sequential text dimension S. Every entry in this volume corresponds to the interaction between the given token and image patch at this position. We make a selection over a slice of this volumn (yellow boxes) and project it onto one of the input modes by summation along the other dimensions (green arrows). This either results in heatmaps over the image dimensions for a text-span selection (top-left) or a heatmap over the text dimension (bottom-left) for a selection of an image region.}
%     \label{fig:fist_vs_second_order_3d}
% \end{figure*}

\section{Stochastic Dominance} \label{sec:stoch_order}

Stochastic dominance defines an order relation between probability distributions based on their cumulatives. \citet{delBarrio} have proposed a significance test building on the principle and \citet{dror} have identified it as being particularly suitable to compare deep neural models.
The test's \(\epsilon\)-parameter is the maximal percentile range where the inferior distribution is allowed to dominate the superior one and Dror \textit{et al.} suggest to set it to \(\epsilon<0.4\). The smaller \(\epsilon\), the stricter the criterion. \(\alpha\) is the significance level.

\section{Integrated Gradients} \label{apx:ig}

We derive \gls{ig} for a model \(f(\mathbf{a}) = s\) with a vector-valued input \(\mathbf{a}\) and a scalar prediction \(s\), e.g. a classification score.
We define the reference input \(\mathbf{r}\), begin from the difference between the two predictions and reformulate it as an integral over the integration variable \(\mathbf{x}\):

\begin{equation} \label{eq:int_grad_diff}
    f(\mathbf{a}) - f(\mathbf{r}) = \int_\mathbf{r}^\mathbf{a} \frac{\partial f(\mathbf{x})}{\partial \mathbf{x}_i} d\mathbf{x}_i
\end{equation}

Again we do not write out sums over double indices.
To solve the resulting line integral, we substitute with the straight line $\mathbf{x}(\alpha) = \mathbf{r} + \alpha (\mathbf{a} - \mathbf{r})$ and pull its derivative $\partial \mathbf{x}(\alpha) / \partial \alpha = (\mathbf{a} - \mathbf{r})$ out of the integral:

\begin{equation} \label{eq:int_grad}
    \int_0^1 \frac{\partial f(\mathbf{x}(\alpha))}{\partial \mathbf{x}_i(\alpha)} \frac{\partial \mathbf{x}_i(\alpha)}{\partial \alpha} d \alpha
    = (\mathbf{a} - \mathbf{r})_i \int_0^{1} \nabla_i f( \mathbf{x}(\alpha)) \, d \alpha
\end{equation}

In practice, we approximate the integral by a sum over \(N\) steps. 
If the reference is uninformative, so that \(f(\mathbf{r})\approx0\), the equality between Eq. \ref{eq:int_grad_diff} and Eq. \ref{eq:int_grad} can be reduced to the final approximation of \gls{ig}:

\begin{equation} \label{eq:ig}
    f(\mathbf{a}) \approx (\mathbf{a} - \mathbf{r})_i \, \frac{1}{N} \sum_{n=1}^N \, \nabla_i f(\mathbf{x}(\alpha_n)),
\end{equation}

which decomposes the model prediction \(f(\mathbf{a})\) into contributions of individual feature \(i\) in \(\mathbf{a}\).

\section{Relation to interactionCAM} \label{apx:gradcam}

Here, we first discuss the relation of integrated gradients \gls{ig} and \textsc{GradCam} and then show how our second-order method can be reduced to the \gls{icam} baseline.

We start from the right-hand-side of Equation \ref{eq:ig}, the final form of \gls{ig}. 
we can reduce these this result further by setting \(N=1\) and using the zero vector as a reference, \(\mathbf{r}=\mathbf{0}\).
These simplifications yield, 

\begin{equation}
    \mathbf{a}_i \nabla_i f(\mathbf{a}),
\end{equation}

which is often referred to as \textit{gradient\(\times\)input} and is the basic form of GradCam. The method typically attributes to deep image representations in CNNs, so that \(\mathbf{a}\) has the dimensions \(C \!\times\! H \!\times\! W\), the number of channels, height and width of the representation.
To reduce attributions to a two-dimensional map, it sums over the channel dimension and applies a relu-activation to the outcome.
The original version also average pools the gradients over the spatial dimensions, however, this is technically not necessary.

As discussed earlier, neither integrated gradients nor \textsc{GradCam} can explain interaction in dual encoder predictions.
Following the logic from above we can, however, reduce our second-order attributions from Eq. \ref{eq:attr_mat} by setting \(N=1\) in the computation of the integrated Jacobians in Eq. \ref{eq:int_jac} and using \(\mathbf{r}_a=\mathbf{r}_b=\mathbf{0}\).
For our attribution matrix from Equation \ref{eq:attr_mat} we then receive the simplified version

\begin{equation}
    \mathbf{a}_i \, \frac{\partial \mathbf{g}_k}{\partial \mathbf{a}_i} \, \frac{\partial \mathbf{h}_k}{\partial \mathbf{b}_j} \, \mathbf{b}_j.
\end{equation}

This simplification could be termed \textit{Jacobians\(\times\)inputs} and is equivalent to the \gls{icam} by \citet{tip}.
Note, however, that setting \(N=1\) is the worst possible approximation to the integrated Jacobians.
Therefore, it is not surprising that empirically this version performs worse than our full attributions.

\section{Interaction LIME} \label{sec:ilime}

We reimplement the \gls{ilime} method proposed by \citet{ilime} that extends the principle of LIME \cite{lime} to dual encoder models with two inputs.\\
The core idea of LIME is to locally approximate the actual model \(f\) around a given input with an interpretable surrogate model \(\varphi\). 
The local neighborhood of the input is approximated by a sample of perturbations.
The surrogate model is typically linear and operates on latent representations of the input. 
Further, there needs to be a mapping from latent representations to input representations, so that we can generate corresponding inputs that the actual model can process.\\
In the image domain latent representations \(\mathbf{z}^a\) are typically binary variables indicating the presence or absence of super pixels in the input.
To enable a direct comparison to our method and the other baselines, we use the vision transformer’s patches as super pixels. 
Analogously, in the text input we define latent representations \(\mathbf{z}^b\) as binary variables indicating the presence of input tokens.
Disabled image patches are replaced with the mean over the image, disabled tokes are replaced with the padding token.\\
The local neighborhood of a given input pair \((\mathbf{a}, \mathbf{b})\) is approximated by sampling \(N\) such latent representations \((\mathbf{z}^a_i, \mathbf{z}^b_i)\) from two Bernoulli distributions.
For the corresponding input perturbations \((\mathbf{a}_i, \mathbf{b}_i)\), we then compute the \gls{clip} scores \(s_i = f(\mathbf{a}_i, \mathbf{b}_i)\) and fit the surrogate model to reproduce these predictions.\\
To account for interactions between the two inputs in dual encoder models, \citet{ilime} propose to use a bilinear form as surrogate model:

\begin{equation}
    \varphi (\mathbf{z}^a, \mathbf{z}^b) = {\mathbf{z}^{a}}^\top \mathbf{W} \mathbf{z}^b + c,  % TODO: transpose z^a
\end{equation} 

with a weight matrix \(\mathbf{W}\) and a scalar bias \(c\), which
is then optimized according to the following MSE objective:

\begin{equation}
    \min_{\mathbf{W}, c}\, \sum_{i=1}^N \,\pi(\mathbf{a}, \mathbf{a}_i, \mathbf{b}, \mathbf{b}_i) \, \Big (f(\mathbf{a}_i, \mathbf{b}_i) - \varphi(\mathbf{z}^a_i, \mathbf{z}^b_i; \mathbf{W}, c) \Big )^2
\end{equation}

Here, \(\pi\) is a function that weights individual neighborhood samples \((\mathbf{a}_i, \mathbf{b}_i)\) according to their similarity to the original input \((\mathbf{a}, \mathbf{b})\).
We use the cosine similarities between perturbed and original captions and image inputs, respectively, and following \citet{ilime}, define the total similarity weight as the average of the caption and image similarity:

\begin{equation}
    \pi(\mathbf{a}, \mathbf{a}_i, \mathbf{b}, \mathbf{b}_i) = \frac{1}{2} \big( \mathbf{g}^\top(\mathbf{a})\, \mathbf{g}(\mathbf{a}_i) + \mathbf{h}^\top(\mathbf{b}) \, \mathbf{h}(\mathbf{b}_i) \big)
\end{equation}

To fit \(\varphi\), we use stochastic gradient descent with a learning rate of \(10^{-2}\) and weight-decay of \(10^{-3}\) over \(N=1000\) samples with Bernoulli drop-out probabilities of \(p=0.3\) for both caption and image representations.
These parameters closely align with \cite{ilime}.
Additionally, we find that scaling the latent representations \(\mathbf{z}^a\) and \(\mathbf{z}^b\) with the square root of the numbers of tokens \(\sqrt{S}\) and image patches \(\sqrt{H\times\, W}\), respectively, helps to stabilize convergence.

Finally, the fitted weight matrix \(\mathbf{W}\) models interactions between image patches and caption tokens. 
Therefore, we can evaluate and visualize it in the same way as our attribution matrices \(\mathbf{A}\).

In Section \ref{sec:attr_eval} we found that \gls{ilime} performs well -- and even slightly better than our method -- on conditional caption attribution.
At the same time its conditional image attributions are not competitive.
Consequently, its grounding ability as evaluated by the PG-metrics is also weak (cf. Table \ref{tab:baselines}).\\
We believe the reason for this imbalance of attribution quality may be due to the different magnitudes in the number of caption tokens and image patches.
While captions typically have \(\sim \! 10\) tokens, image representations in ViT-B-32 architectures consist of \(\sim\!200\) patches.
Therefore, the ratio of the number of samples \(N\) and tokens is much better than for image patches and the surrogate model \(\varphi\) might be able estimate their importances better.

Overall, we find that the optimization of \gls{ilime} is quite sensitive to hyper-parameter choices and requires extensive tuning to find a setting that leads to stable convergence.
In contrast, our method does not require additional optimization and involves no hyper-parameters except the number of integration steps \(N\), whose increase must, however, improve attributions due to Equation \ref{eq:int_jac}.

\section{Implementation Details} \label{apx:implementation}
For the implement of our method, we make use of the auto-differentiation framework in the PyTorch package.
For a give input \(\mathbf{x}(\alpha_n)\), \(\mathbf{g}(\mathbf{x}(\alpha_n))\) is the forward pass through the encoder \(\mathbf{g}\), and the Jacobian \(\partial \mathbf{g}_k(\mathbf{x}(\alpha_n)) / \partial \mathbf{x}_i\) is the corresponding backward pass.
For an efficient computation of all \(N\) interpolation steps in Eq. \ref{eq:int_jac}, we can batch forward and backward passes since individual interpolations are independent of another.\\
In practice, we attribute to intermediate representations, thus, the interpolations in Eq. \ref{eq:lines} are between latent representations of the references and inputs.
We use PyTorch \textit{hooks} to compute these interpolations during the forward pass. Algorithm \ref{alg:pseudocode} sketches PyTorch-like pseudo-code of the implementation.

The application of our method to a different model or architecture only requires the implementation of a single forward hook.
Registering hooks into models is a standard feature in auto-differentiation frameworks and does not require any modification of the given model's original code.
The remaining steps to generate our attributions are differentiation through standard backpropagation and, finally, simple matrix multiplication to compute Eq. \ref{eq:attr_mat}.

\section{Computational Complexity}
Since the computation of the interpolated inputs \(\mathbf{x}(\alpha_n)\) can be performed in parallel, \(N\) is a constant with regard to time complexity.
To build the full Jacobians of the encoders, however, we need to compute a separate backward pass for each output dimension, because auto-differentiation can only compute backward passes for scalar-valued outputs.
Time complexity is dominated by this aspect and is thus on the order of \(O(D)\), with \(D\) being the embedding dimensionality of the output.
The \texttt{intergrated\_jacobian} method in Algorithm \ref{alg:pseudocode} sketches this computation.\\
Due to the fact that we typically attribute to intermediate representations, however, we do not need to compute full backward passes.
Backpropagation can be stopped once it reaches the representation we attribute to, which results in this operation to be cheaper the deeper the representation of interest is.
I.e. attributing to layer eleven is cheaper than attributing to layer five.\\
After building the Jacobians, the final attributions are computed through the matrix multiplications in Eq. \ref{eq:attr_mat}, which is outlined at the bottom of Algorithm \ref{alg:pseudocode}. 
Its computation time is negligible when calculated on GPU, but can substantially add to the total time when performed on CPU.

Space-wise, our method requires storing two Jacobians with the dimensions \(D \times (D \times S)\), and \(D \times (D \times H \times W)\), since input/intermediate representations are still sequential (\(S\)) on the text side and patch-based (\(H \times W\)) on the image side.
Thus, memory consumption scales quadratically on the order of \(O(D^2)\), and we require large VRAM to handle the computation efficiently on GPU.

In contrast, first-order methods only require backpropagation of a single scalar output value, i.e. the similarity score, whose result is a gradient vector as opposed to a Jacobian matrix.
Hence, the cost of obtaining our second-order interaction attribution is the computation and handling of these Jacobians, which is substantially more expensive but enables a different level of insight into models that is not accessible through first-order methods. 

\begin{algorithm}
\begin{minted}{python}
from torch import Tensor, arange, stack, autograd
import ExplainableCLIP, image_preparation, tokenize

model         = ExplainableCLIP(...)
image_input   = image_preparation(load_image("path/to/image.png"))
caption_input = tokenize("some caption describing the image")

# Equations 6 and 7
def interpoloate(x: Tensor, ref: Tensor, n_steps: int):
    '''
    Compute n_steps linear interpolations between a reference ref and input x.
    '''
    step = 1 / n_steps
    alphas = arange(1, 0, step)  # interpolation coefficients
    x_interp = ref + alphas * (x - ref)  # interpolated representations
    return x_interp
    
# Equation 9
def integrated_jacobian(embedding: Tensor, intermediate: Tensor):
    '''
    Compute the integrated Jacobian for an embedding w.r.t. an 
    intermediate representation.
    '''
    gradients = []
    for dim in range(embedding.size(0)):
        grad_d = autograd.grad(embedding[dim], intermediate)
        gradients.append(grad_d)
    jacobians = stack(gradients)
    # Integration over interpolations stacked along the first dimension
    int_jacobian = jacobians.sum(dim=0)
    return int_jacobian

# place hooks in the model to compute interpolations (not actual hook syntax)
image_hook   = model.register_hook(interpolate, image_layer, image_reference,   n_steps)
caption_hook = model.register_hook(interpolate, text_layer,  caption_reference, n_steps)

# Compute embeddings and retrieve intermediate representations from hooks
image_embedding   = model.encode_image(image_input)
caption_embedding = model.encode_caption(caption_input)
image_inter, image_ref_inter     = image_hook.get_intermediate_representation()
caption_inter, caption_ref_inter = caption_hook.get_intermediate_representation()

# Equation 10
image_jacobian   = integrated_jacobian(image_embedding,   image_inter)
caption_jacobian = integrated_jacobian(caption_embedding, caption_inter)
# Matrix multiplication
JJ = caption_jacobian.T @ image_jacobian 
image_delta   = image_inter   - image_ref_inter
caption_delta = caption_inter - caption_ref_inter
# Element-wise multiplication with broadcasting
attributions = caption_delta * JJ * image_delta
\end{minted}
\caption{PyTorch-like pseudocode sketching the computation of our attributions. The syntax is simplified and not consistent. For a fully functional implementation, please refer to our GitHub repository. Comments in the pseudocode refer to the corresponding equations in Section \ref{sec:method}.}
\label{alg:pseudocode}
\end{algorithm}

\end{document}